\documentclass[letterpaper,10pt,journal]{IEEEtran}
\IEEEoverridecommandlockouts

\usepackage[colorlinks,linkcolor=black,citecolor=black,urlcolor=black]{hyperref}
\newcommand\rurl[1]{%
\texttt{\href{http://#1}{\nolinkurl{#1}}}%
}

\usepackage[acronym]{glossaries}
\newacronym{slam}{SLAM}{simultaneous localisation and mapping}
\newacronym{sdk}{SDK}{software development kit}

\usepackage{fontawesome}
\usepackage{pifont}

\usepackage{amsfonts}

\usepackage{cite}

\usepackage[labelformat=simple]{subcaption}

\usepackage{algorithm}
\usepackage{algpseudocode}

\usepackage{cleveref}
\crefname{table}{Tab.}{Tabs.}
\crefname{figure}{Fig.}{Figs.}
\crefname{section}{Sec.}{Secs.}
\crefname{equation}{Eq.}{Eqs.}
\crefname{listing}{Lst.}{Lsts.}

\usepackage{graphicx}
\usepackage{cuted}
\usepackage{stackengine}

\usepackage[binary-units]{siunitx}
\usepackage{threeparttable}
\usepackage{booktabs}

\usepackage{tikz}
\usetikzlibrary{shapes,arrows}

\usepackage{multicol}

\usepackage{xcolor}

\usepackage{listings}

\usepackage{crossreftools}

\usepackage{todonotes}

\usepackage[T1]{fontenc} 
\usepackage{combelow}

\input{review.cls}

\begin{document}

\replyLetter


\bstctlcite{IEEEexample:BSTcontrol}

\title{
\huge \bf
\textit{OORD}: The Oxford Offroad Radar Dataset
}
\author{Matthew Gadd, Daniele De Martini, Oliver Bartlett, Paul Murcutt, Matt Towlson, Matthew Widojo,\\Valentina Mu\cb{s}at, Luke Robinson, Efimia Panagiotaki, Georgi Pramatarov, Marc Alexander K\"{u}hn,\\Letizia Marchegiani, Paul Newman, and Lars Kunze\\[5pt]
Oxford Robotics Institute (ORI), University of Oxford, UK\\[5pt]
\faEnvelope~\texttt{mattgadd@robots.ox.ac.uk}
\faDatabase~\rurl{oxford-robotics-institute.github.io/oord-dataset}
\faSmileO~\rurl{huggingface.co/mttgdd/oord-models}
\faGithub~\rurl{github.com/mttgdd/oord-dataset}
}
\maketitle

\begin{strip}
\centering
\vspace{-30pt}
\includegraphics[width=\textwidth]{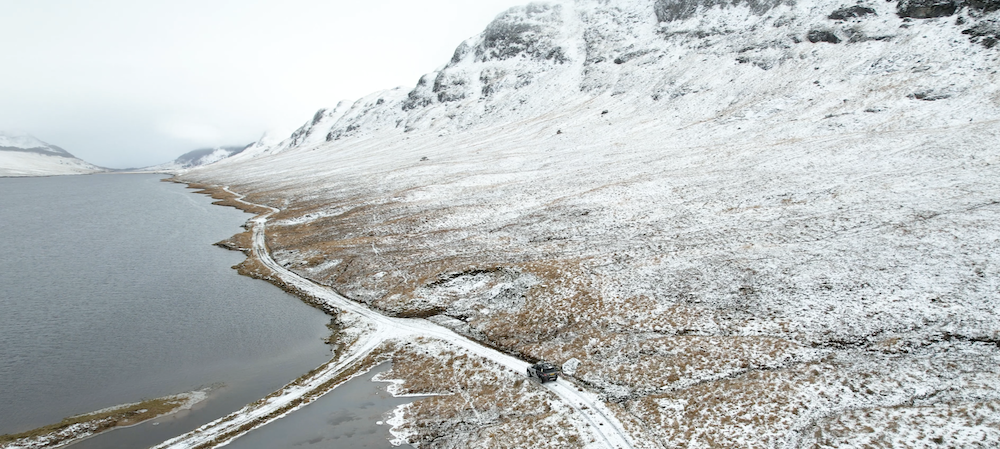}
\captionof{figure}{
Our data collection site is in the Scottish Highlands, with comprehensive coverage of \textit{Ardverikie Estate}, close to the historic boundary between \textit{Lochaber} and \textit{Badenoch}.
This image is taken on
\textit{Lochan na h-Earba} from our \ref{DA1} and \ref{DA2}\fixM{}{4.4.2} datasets (\cref{sec:twolochs,sec:routes_summary}), showing unpaved terrain over, uneven landscape next to the vehicle, and the inclement weather.
Images, GPS traces, and example radar scans for more specific areas of \textit{Ardverikie Estate} are provided in~\cref{fig:dataset_overview_a,fig:dataset_overview_b}.
\label{fig:lochannahearba}
}
\end{strip}

\begin{abstract}
There is a growing academic interest as well as commercial exploitation of millimetre-wave scanning radar for autonomous vehicle localisation and scene understanding.
Although several datasets to support this research area have been released, they are primarily focused on urban or semi-urban environments~\cite{yu2020bdd100k,mei2022waymo,panagiotaki2024robotcycle,suleymanov2021oxford}.
Nevertheless, rugged offroad deployments are important application areas which also present unique challenges and opportunities for this sensor technology.
Therefore, the \textit{Oxford Offroad Radar Dataset (OORD)} presents data collected in the rugged Scottish highlands in extreme weather.
The radar data we offer to the community are accompanied by GPS/INS reference -- to further stimulate research in radar place recognition.
In total we release over \SI{90}{\gibi\byte} of radar scans as well as GPS and IMU readings by driving a diverse set of four routes over $11$ forays, totalling approximately \SI{154}{\kilo\metre} of rugged driving.
This is an area increasingly explored in literature, and we therefore present and release examples of recent open-sourced radar place recognition systems and their performance on our dataset.
This includes a learned neural network, the weights of which we also release.
The data and tools are made freely available to the community at \rurl{oxford-robotics-institute.github.io/oord-dataset}
\end{abstract}

\section{Introduction}
\label{sec:introduction}

Compared to cameras and LiDARs, radar systems offer distinct advantages, including extended-range perception and resilience to lightning and various weather conditions.
\fixM{
In more detail, scanning radar of the type we use in this work does not give 3D information, and is suspect to complex measurement artefacts (speckle noise, ghost reflections, etc), but is extremely robust to atmospheric effects and low-light.
Cameras are information-rich and highly discriminative, but are highly susceptible to appearance variation, e.g. low-light.
LiDARs give detailed scans, often in 3D, and emit their own light so are suitable for low-light operation, but are often limited in range and are nevertheless susceptible to particulate effects, e.g. snow or dust.
}{1.1}
Full reviews of millimetre-wave radar applications in autonomous vehicle tasks are available in e.g.~\cite{harlow2023new,venon2022millimeter}.
In brief, scanning radar-based motion estimation~\cite{cen2018precise,barnes2019masking,park2020pharao,kung2021normal,burnett2021we,adolfsson2021cfear,burnett2021radar,aldera2022goes}, pose estimation~\cite{barnes2020under,yin2020radar,yin2021rall}, place recognition~\cite{kim2020mulran,suaftescu2020kidnapped,barnes2020under,wang2021radarloc,yin2021radar,komorowski2021large,gadd2021contrastive,lu2022one,cait2022autoplace,jang2023raplace,yuan2023iros}, and full \gls{slam}~\cite{hong2020radarslam,wang2022maroam} are increasingly popular in the literature.
This dataset is designed to support that work, and to prompt an extension into naturalistic, rugged environments, which have not yet been explored.

Natural environments are important because off-road autonomous vehicles have many essential applications across industries operating on rugged terrain.
This technology is essential for tasks such as agricultural operations, search and rescue missions, and exploration in remote areas, where human intervention may be impractical or hazardous.
Also, autonomous hauling or resource extraction is important in mining, autonomous tractors for planting and harvesting can optimise crop yields, tree planting and logging can reduce environmental impact and risk to workers, autonomous grading and excavation can enhance efficiency and safety in construction, search and rescue missions can leverage autonomous drones to locate missing persons in hazardous locations, and autonomous vehicles can be of great use in environmental monitoring and disaster response.

In deploying autonomous machines to these sites, place recognition is vital for navigation and localisation.
For this, more robust sensors are crucial to enhancing place recognition accuracy and reliability.
Radar in particular can handle challenging conditions like low-light environments.

We therefore present an offroad radar dataset with a focus on relocalisation, with~\cref{fig:lochannahearba} showing an example location from our collection site.
Our contributions are as follows:
\begin{enumerate}
\item \textbf{Offroad scanning radar data} We release the first radar dataset focusing on off-road, difficult terrain and naturalistic environments,
\item \textbf{Poor weather and lighthing conditions} including collection after thick snowfall and in total darkness (being in the wilderness),
\item \textbf{Place recognition ground truth} Our raw data is referenced against good GPS as a ground truth for the place recognition task,
\item \textbf{Open-sourced radar place recognition implementations} For the first time, we release open-source a set of trained weights for a deep neural network which solves the radar place recognition,
\item \textbf{Software development kit} We release software tools for making the use of this dataset in some pre-existing, non-learned open-source radar place recognition implementations easy, as well as pipelines to accelerate new learning solutions.
\end{enumerate}

In~\cref{sec:routes} we describe and justify the choice of collection sites.
\cref{sec:dataset} describes the data collection platform, sensor format including radar as well as the GPS/INS synchronised with this main sensor, and the download data structure for how to interact with the dataset.
\cref{sec:tools} presents several useful software tools which make it easy to apply this data, e.g. with common performance metrics and data formats. 
Finally,~\cref{sec:examples} shows examples based off of the tools in \cref{sec:tools} for several publicly available place recognition algorithms as tested against this dataset.

\begin{figure*}[!h]
\centering

\begin{subfigure}{\textwidth}\centering

\begin{subfigure}{0.25\textwidth}\centering
\stackunder{\includegraphics[width=\linewidth]{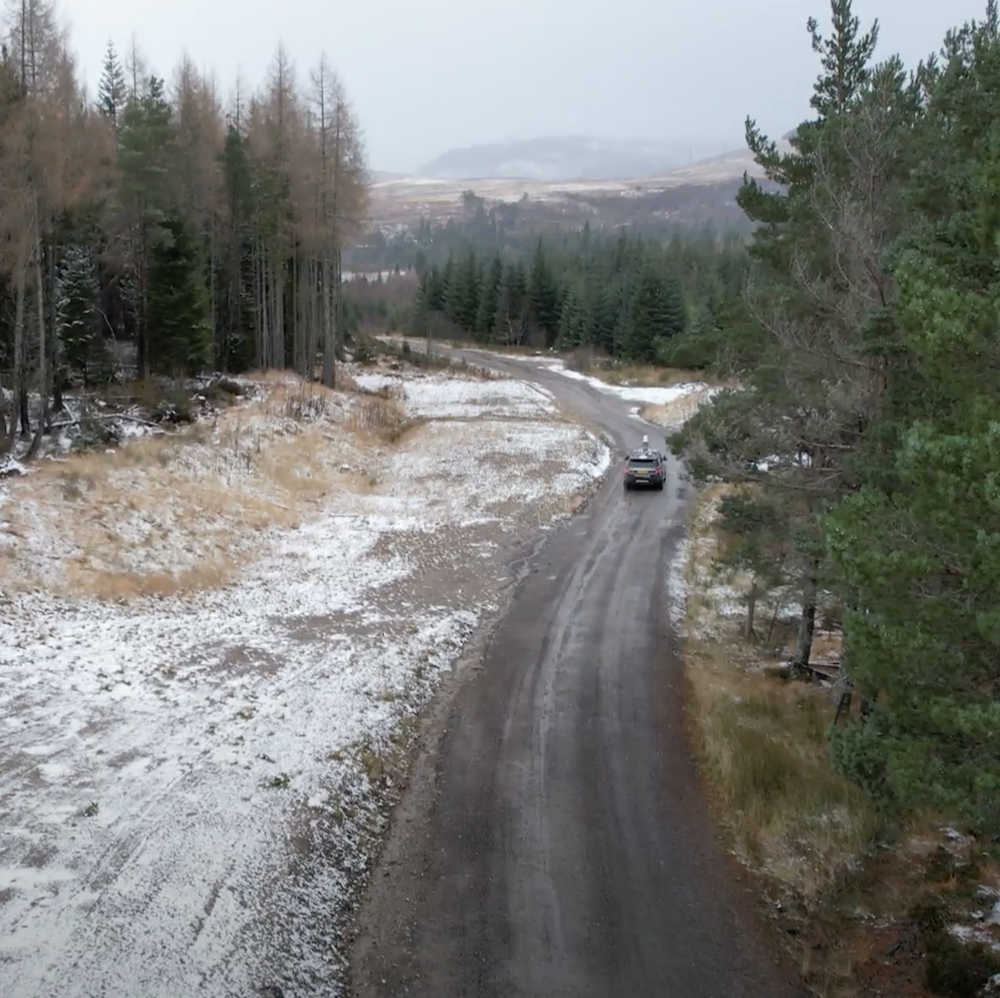}}{\fixM{Drone image of route.}{}}
\end{subfigure}
\begin{subfigure}{0.25\textwidth}\centering
\stackunder{\includegraphics[width=\linewidth]{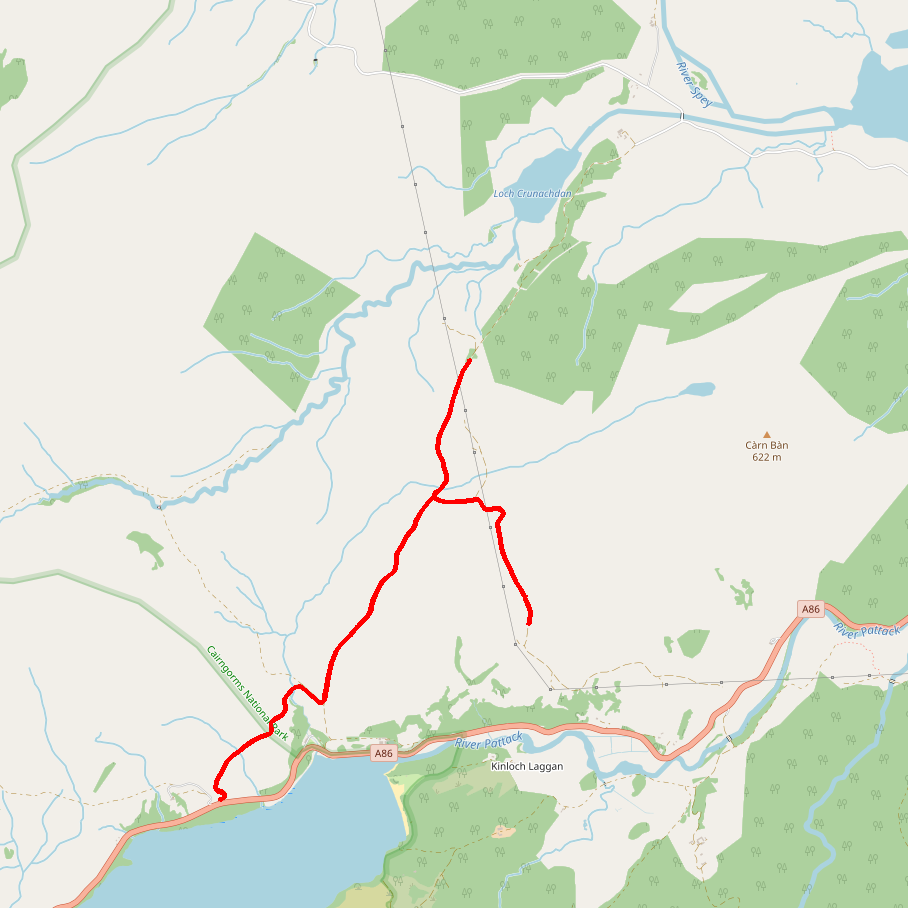}}{\fixM{GPS trace (red).}{}}
\end{subfigure}
\begin{subfigure}{0.25\textwidth}\centering
\stackunder{\includegraphics[width=\linewidth]{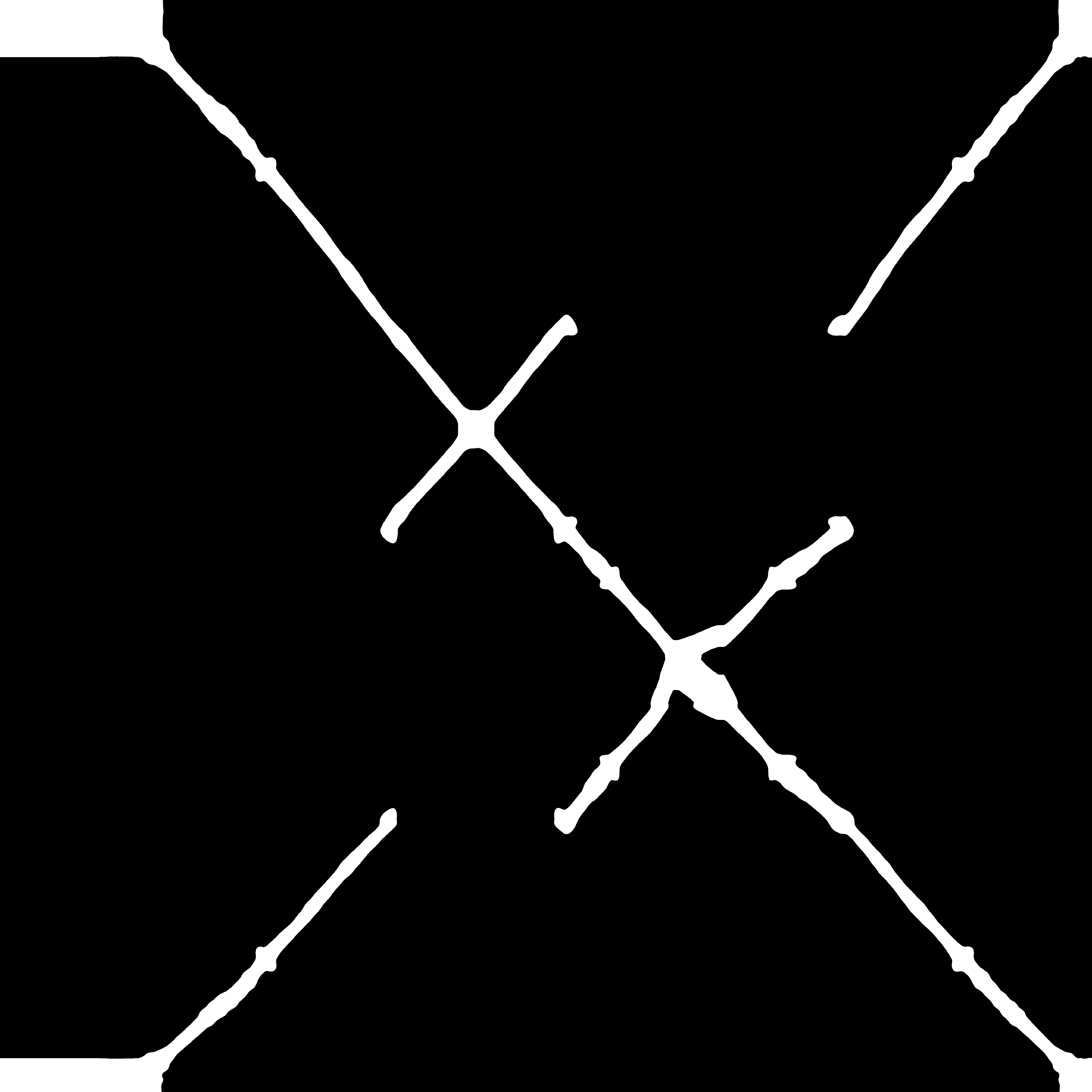}}{\fixM{Ground truth matrix.}{}\label{fig:bellmouth_tp}}
\end{subfigure}

\vspace{5pt}

\begin{subfigure}{0.1875\textwidth}\centering
\stackunder{\includegraphics[width=\linewidth]{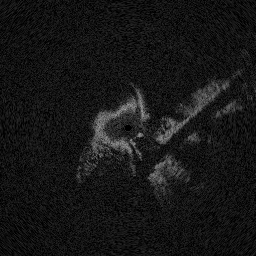}}{\fixM{Example scan \#1.}{}}
\end{subfigure}
\begin{subfigure}{0.1875\textwidth}\centering
\stackunder{\includegraphics[width=\linewidth]{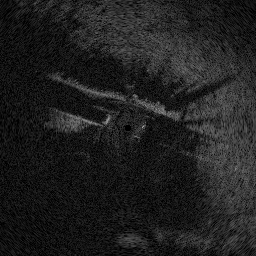}}{\fixM{Example scan \#2.}{}}
\end{subfigure}
\begin{subfigure}{0.1875\textwidth}\centering
\stackunder{\includegraphics[width=\linewidth]{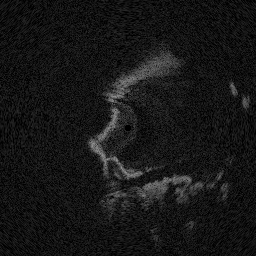}}{\fixM{Example scan \#3.}{}}
\end{subfigure}
\begin{subfigure}{0.1875\textwidth}\centering
\stackunder{\includegraphics[width=\linewidth]{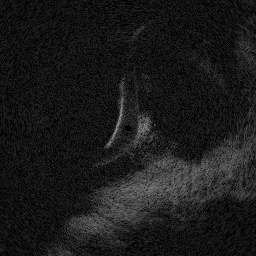}}{\fixM{Example scan \#4.}{4.2.1}}
\end{subfigure}

\caption{
The \texttt{bellmouth} dataset is on gravel tracks ascending and descending a mountain trail, with the descent along the same path but in the opposite direction.
\label{fig:bellmouth_overview}
}
\end{subfigure}

\vspace{5pt}

\begin{subfigure}{\textwidth}\centering

\begin{subfigure}{0.25\textwidth}\centering
\stackunder{\includegraphics[width=\linewidth]{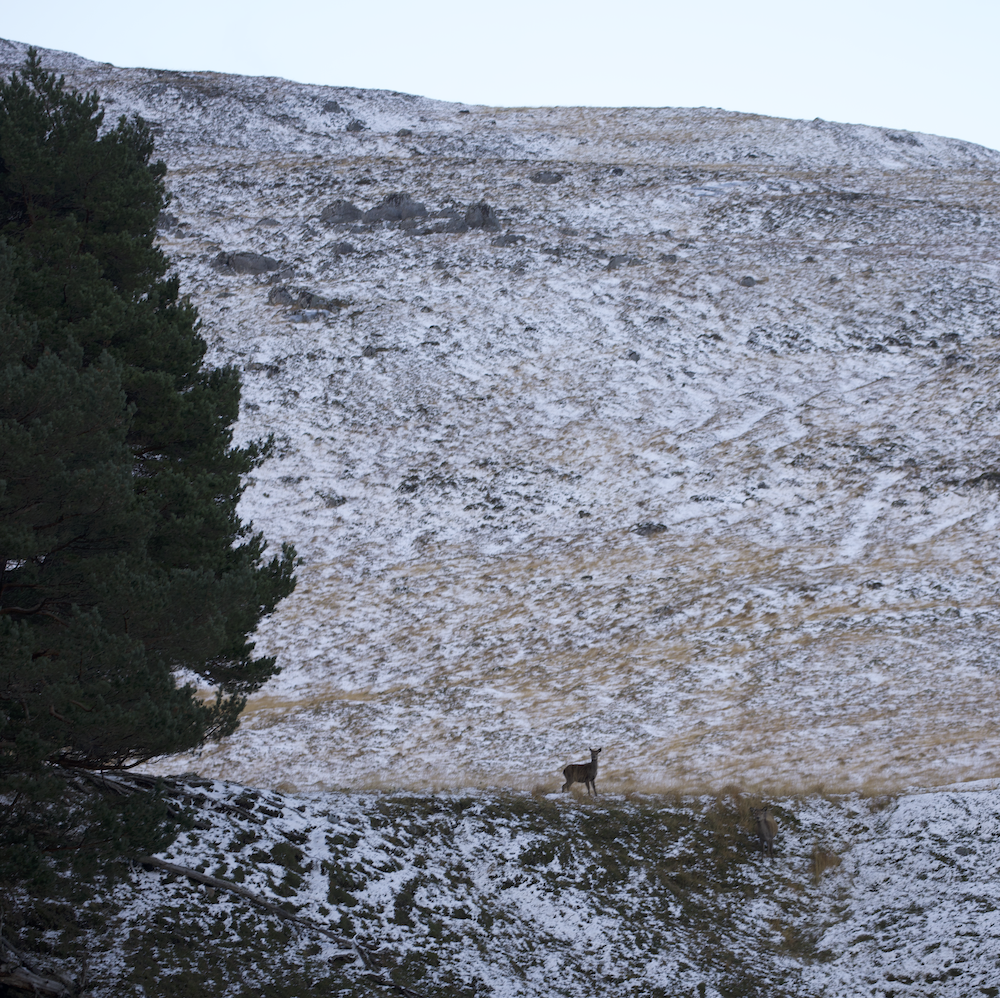}}{\fixM{DSLR image from route.}{}}
\end{subfigure}
\begin{subfigure}{0.25\textwidth}\centering
\stackunder{\includegraphics[width=\linewidth]{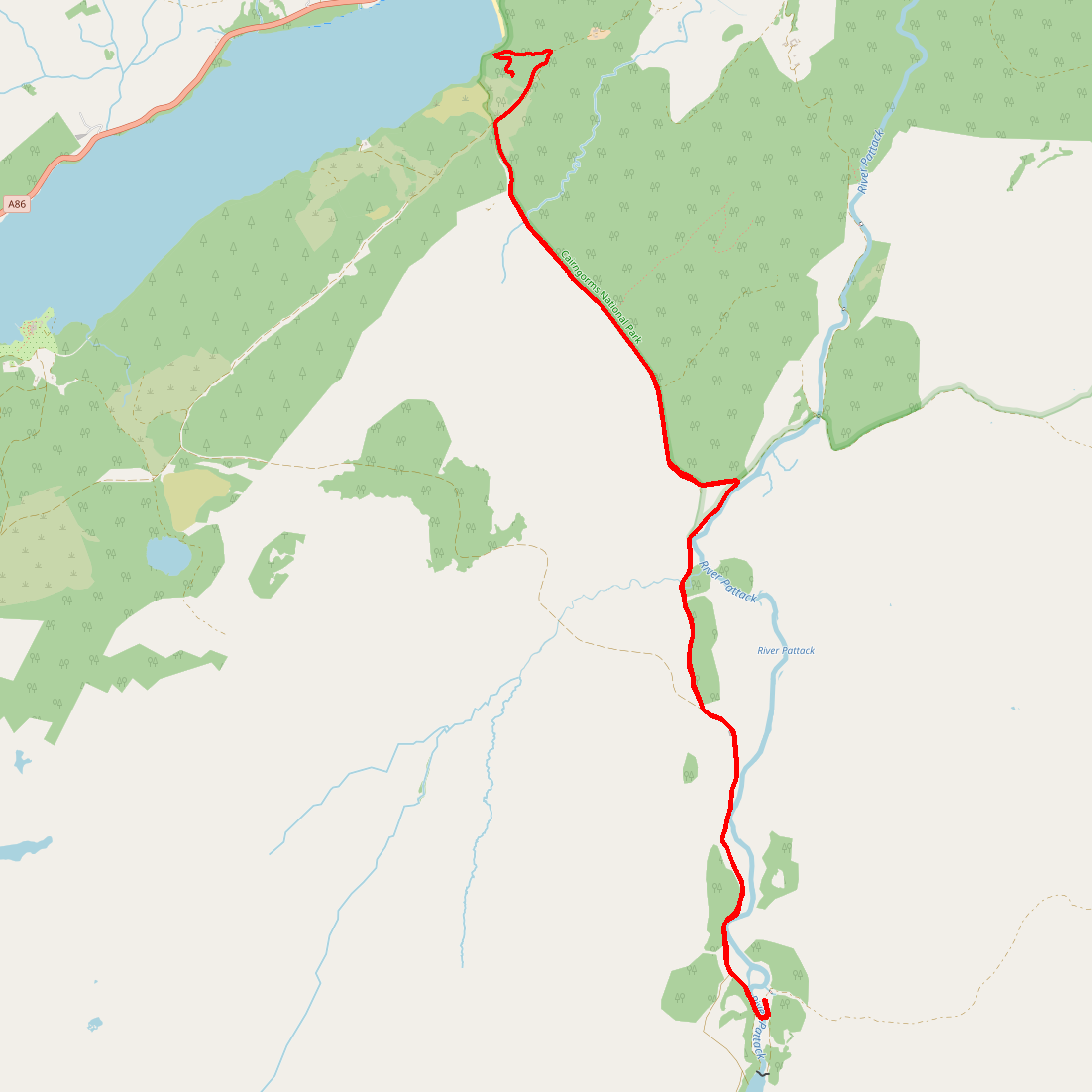}}{\fixM{GPS trace (red).}{}}
\end{subfigure}
\begin{subfigure}{0.25\textwidth}\centering
\stackunder{\includegraphics[width=\linewidth]{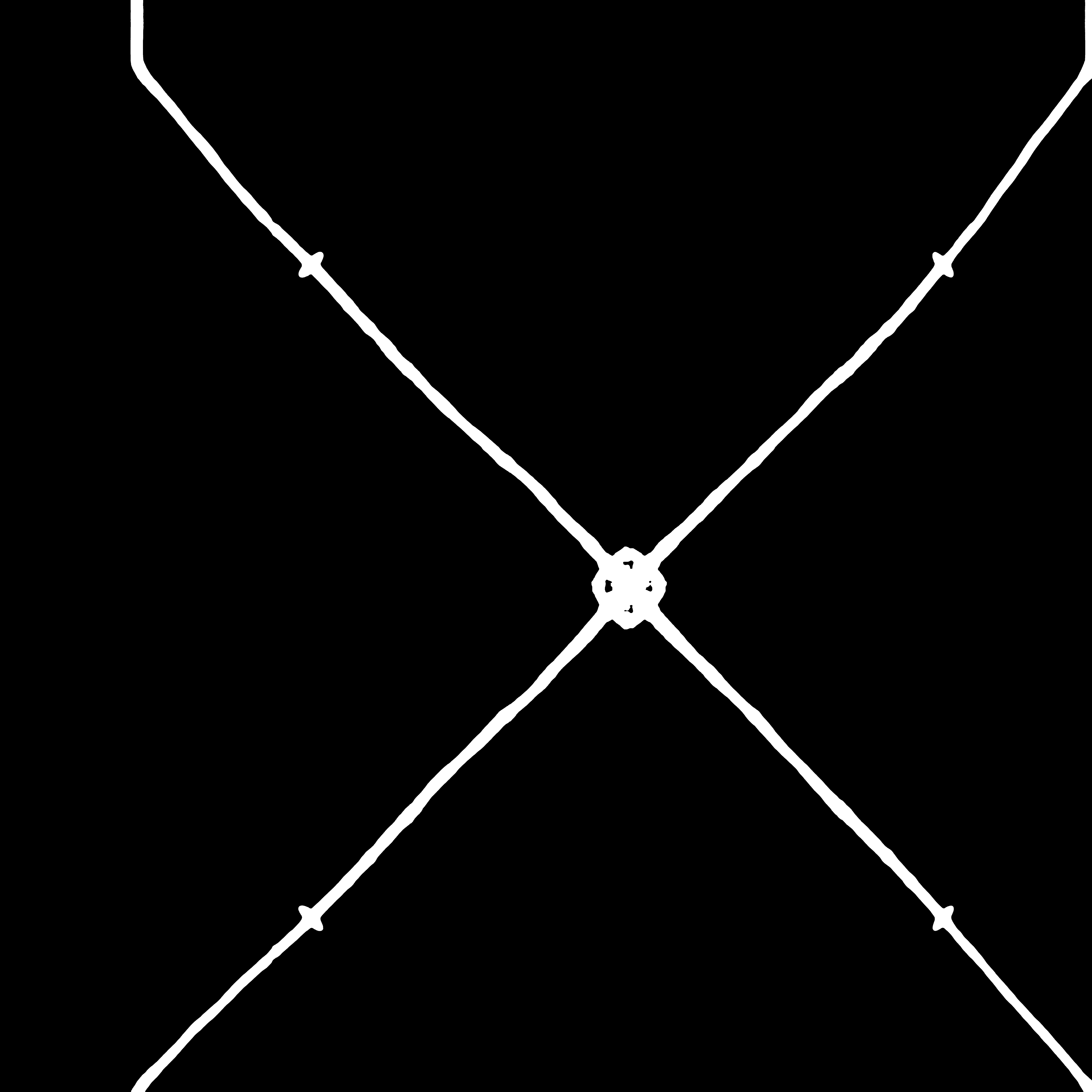}}{\fixM{Ground truth matrix.}{}\label{fig:hydro_tp}}
\end{subfigure}

\vspace{5pt}

\begin{subfigure}{0.1875\textwidth}\centering
\stackunder{\includegraphics[width=\linewidth]{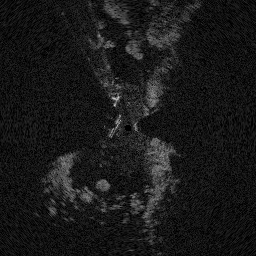}}{\fixM{Example scan \#1.}{}}
\end{subfigure}
\begin{subfigure}{0.1875\textwidth}\centering
\stackunder{\includegraphics[width=\linewidth]{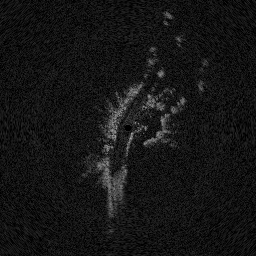}}{\fixM{Example scan \#2.}{}}
\end{subfigure}
\begin{subfigure}{0.1875\textwidth}\centering
\stackunder{\includegraphics[width=\linewidth]{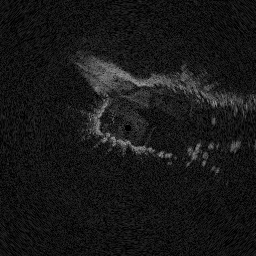}}{\fixM{Example scan \#3.}{}}
\end{subfigure}
\begin{subfigure}{0.1875\textwidth}\centering
\stackunder{\includegraphics[width=\linewidth]{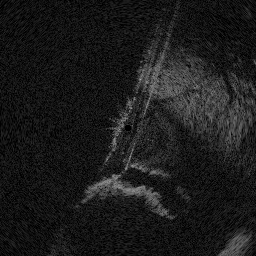}}{\fixM{Example scan \#4.}{4.2.2}}
\end{subfigure}

\caption{
The \texttt{hydro} dataset ascends to the highest elevation of all of our routes, and descends along the same track.
Much of the route is along the \textit{River Pattack}.
\label{fig:hydro_overview}}
\end{subfigure}

\caption{
The \texttt{bellmouth} and \texttt{hydro} datasets, with GPS traces, trajectory-to-trajectory ground truth GPS matrices, and some sample radar scans.
}
\label{fig:dataset_overview_a}
\end{figure*}

\begin{figure*}[!h]
\centering

\begin{subfigure}{\textwidth}\centering

\begin{subfigure}{0.25\textwidth}\centering
\stackunder{\includegraphics[width=\linewidth]{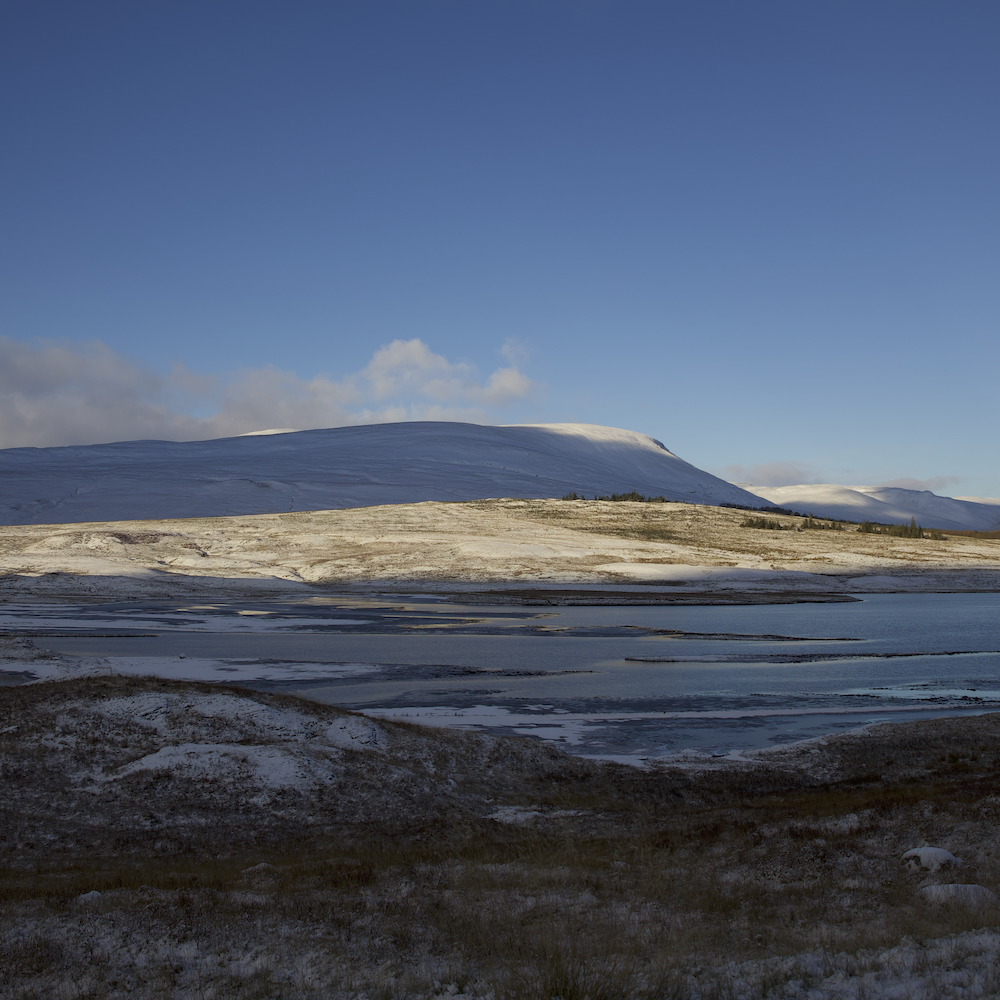}}{\fixM{DSLR image from route.}{}}
\end{subfigure}
\begin{subfigure}{0.25\textwidth}\centering
\stackunder{\includegraphics[width=\linewidth]{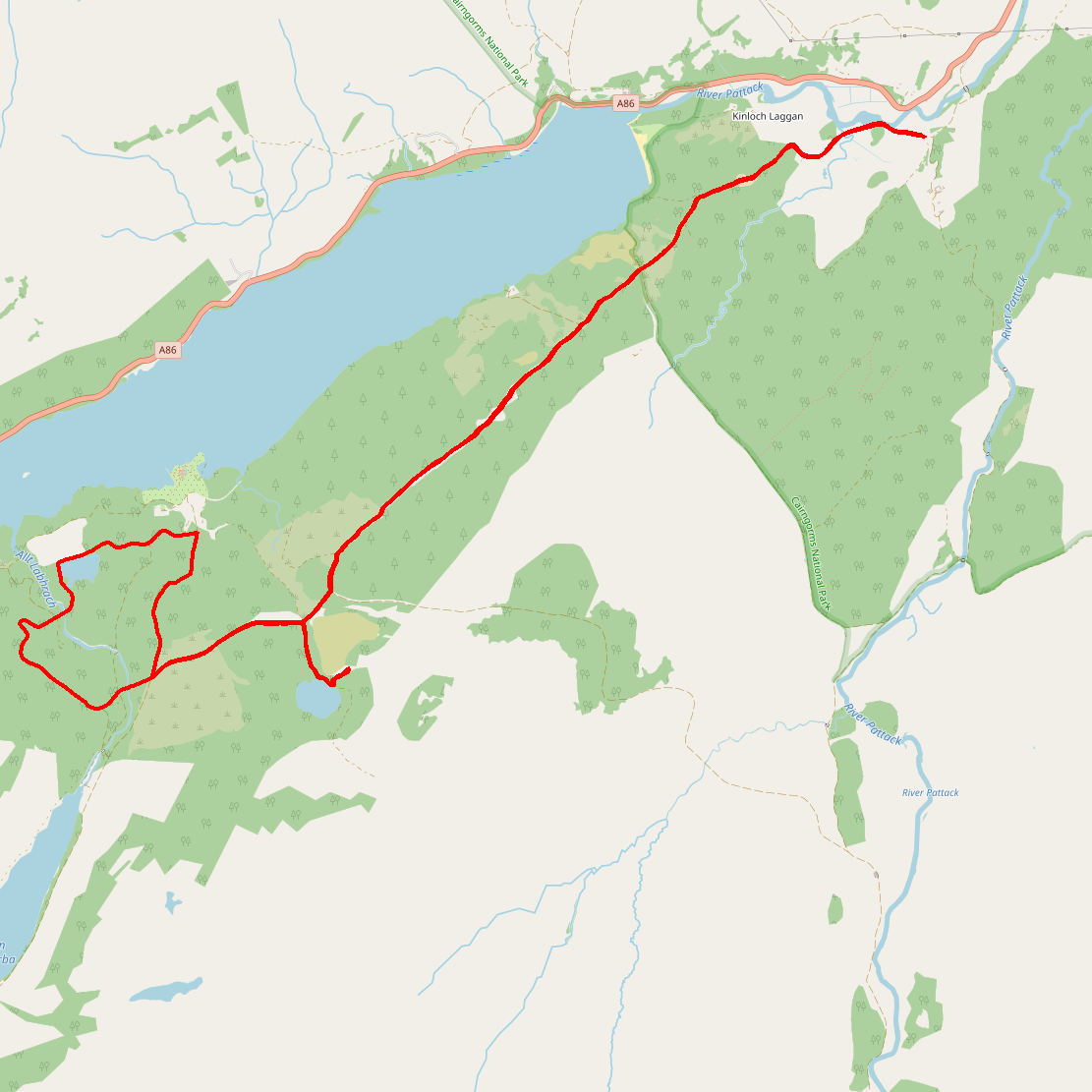}}{\fixM{GPS trace (red).}{}}
\end{subfigure}
\begin{subfigure}{0.25\textwidth}\centering
\stackunder{\includegraphics[width=\linewidth]{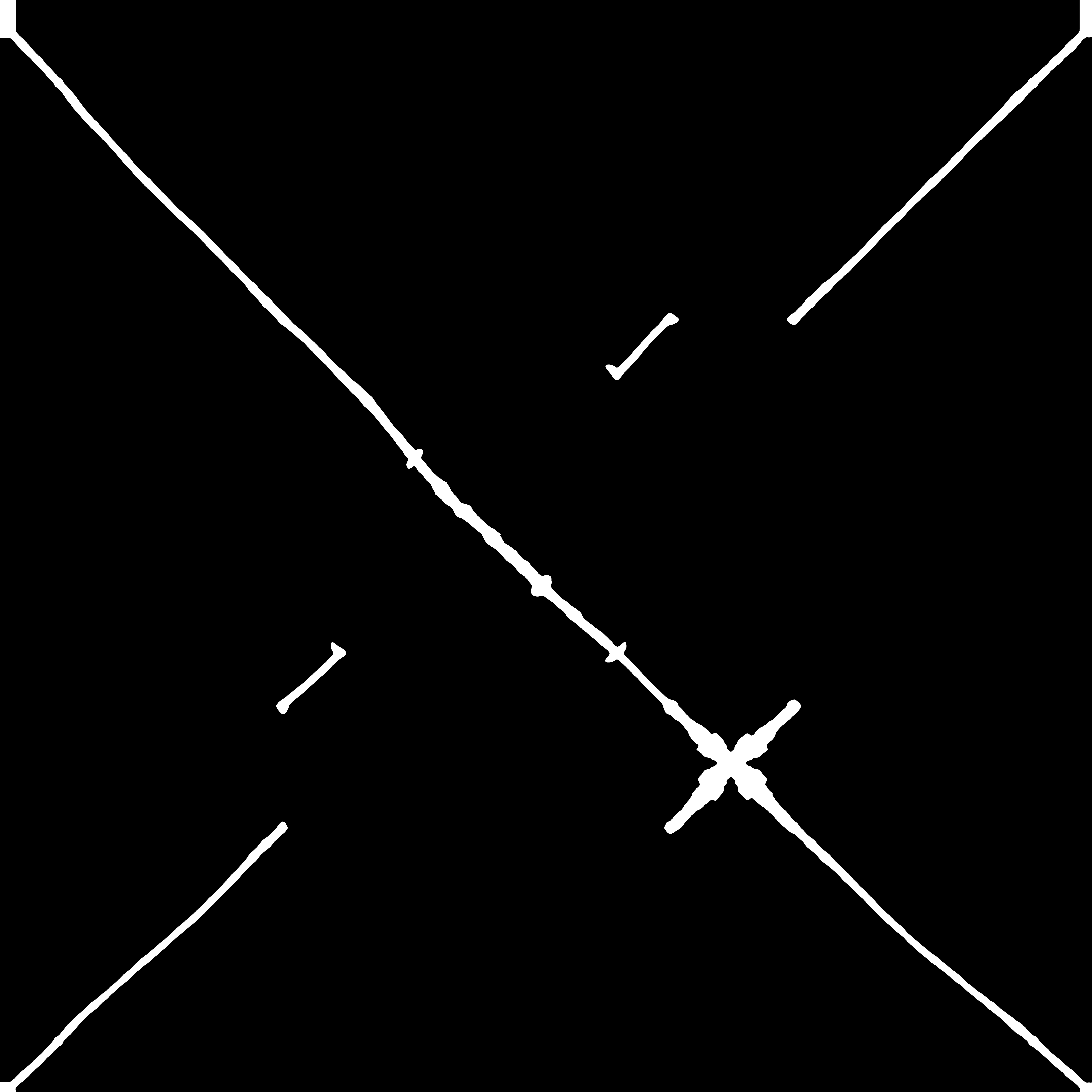}}{\fixM{Ground truth matrix.}{}\label{fig:maree_tp}}
\end{subfigure}

\vspace{5pt}

\begin{subfigure}{0.1875\textwidth}\centering
\stackunder{\includegraphics[width=\linewidth]{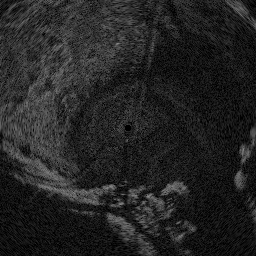}}{\fixM{Example scan \#1.}{}}
\end{subfigure}
\begin{subfigure}{0.1875\textwidth}\centering
\stackunder{\includegraphics[width=\linewidth]{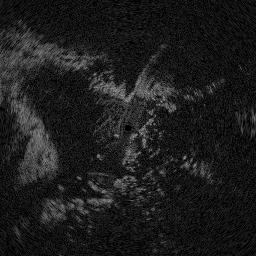}}{\fixM{Example scan \#2.}{}}
\end{subfigure}
\begin{subfigure}{0.1875\textwidth}\centering
\stackunder{\includegraphics[width=\linewidth]{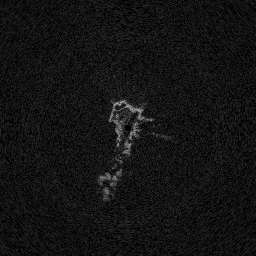}}{\fixM{Example scan \#3.}{}}
\end{subfigure}
\begin{subfigure}{0.1875\textwidth}\centering
\stackunder{\includegraphics[width=\linewidth]{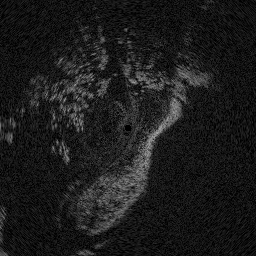}}{\fixM{Example scan \#4.}{4.2.3}}
\end{subfigure}

\caption{
The \texttt{maree} dataset shares outgoing and incomings routes in opposite d4.2.4irections, with a single-direction loop midway.
\label{fig:maree_overview}}
\end{subfigure}

\vspace{5pt}

\begin{subfigure}{\textwidth}\centering

\begin{subfigure}{0.25\textwidth}\centering
\stackunder{\includegraphics[width=\linewidth]{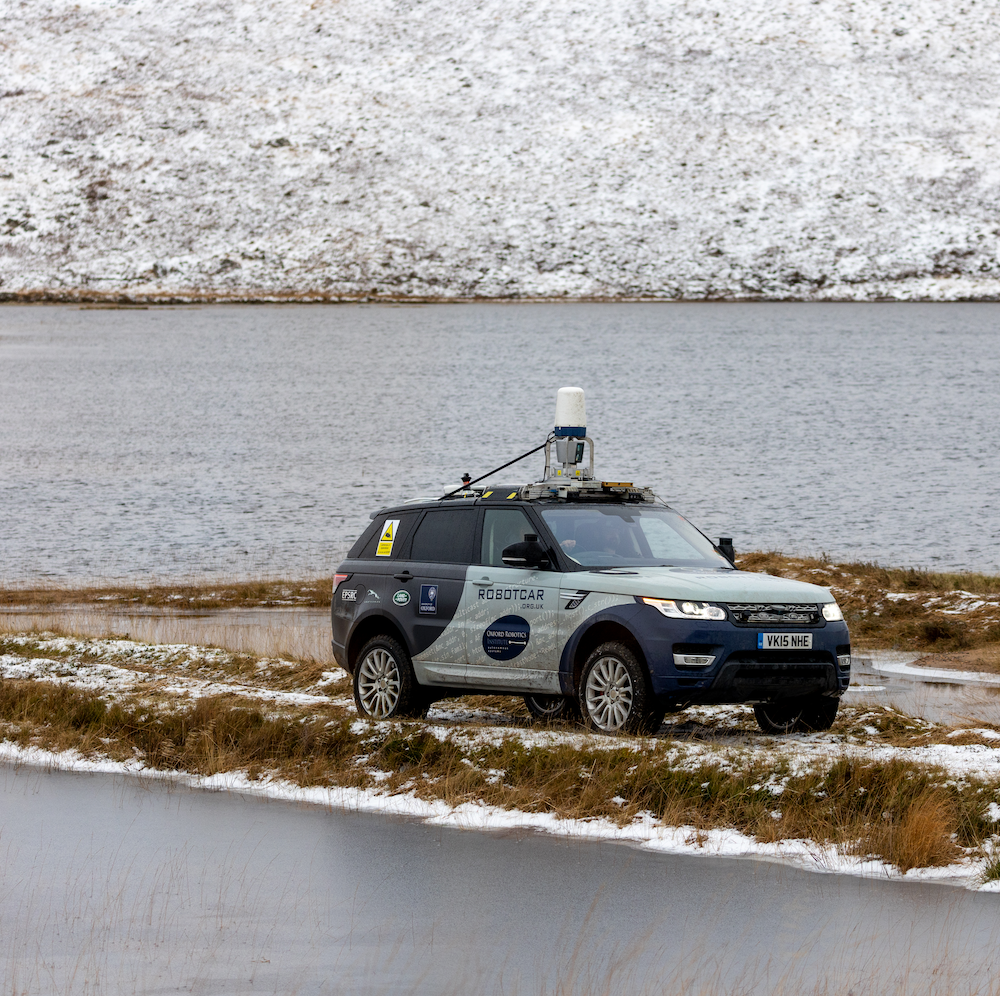}}{\fixM{DSLR image from route.}{}}
\end{subfigure}
\begin{subfigure}{0.25\textwidth}\centering
\stackunder{\includegraphics[width=\linewidth]{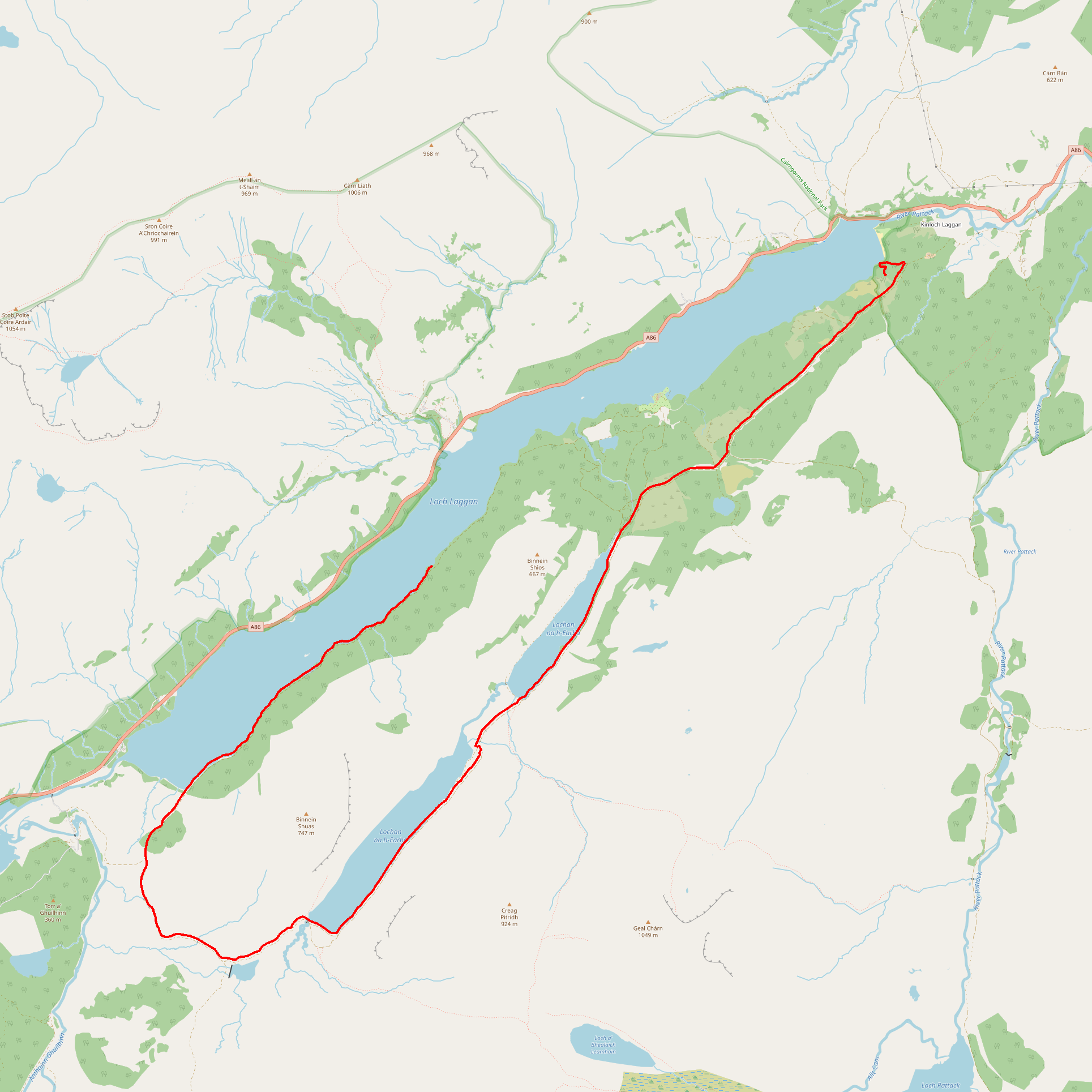}}{\fixM{GPS trace (red).}{}}
\end{subfigure}
\begin{subfigure}{0.25\textwidth}\centering
\stackunder{\includegraphics[width=\linewidth]{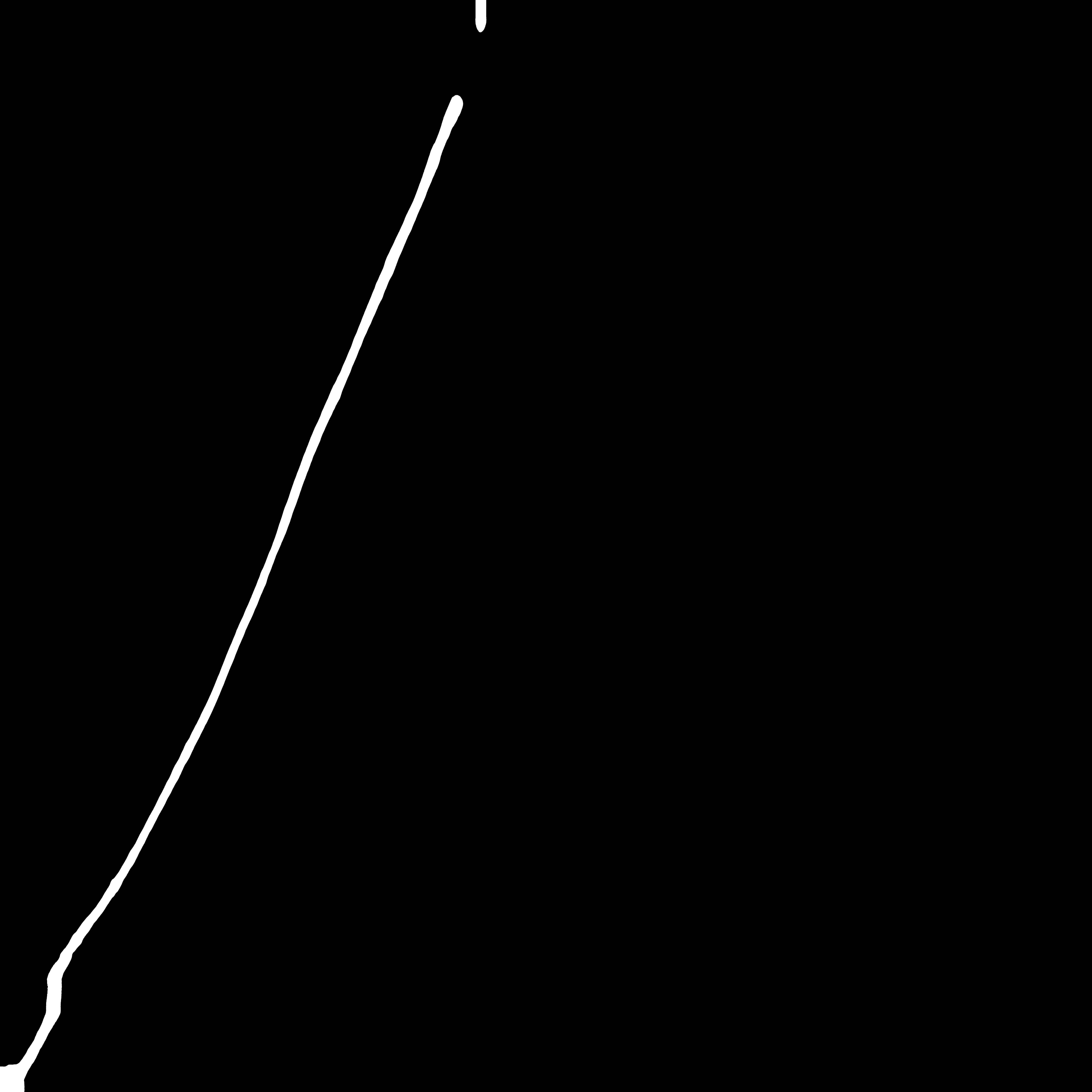}}{\fixM{Ground truth matrix.}{}\label{fig:twolochs_tp}}
\end{subfigure}

\vspace{5pt}

\begin{subfigure}{0.1875\textwidth}\centering
\stackunder{\includegraphics[width=\linewidth]{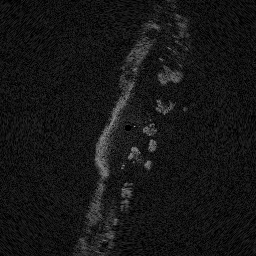}}{\fixM{Example scan \#1.}{}}
\end{subfigure}
\begin{subfigure}{0.1875\textwidth}\centering
\stackunder{\includegraphics[width=\linewidth]{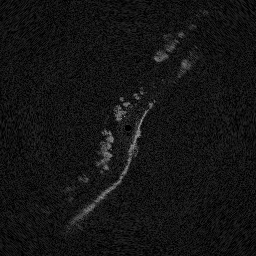}}{\fixM{Example scan \#2.}{}}
\end{subfigure}
\begin{subfigure}{0.1875\textwidth}\centering
\stackunder{\includegraphics[width=\linewidth]{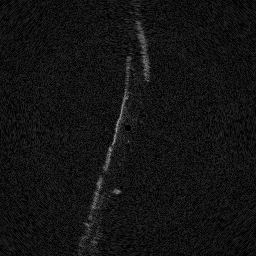}}{\fixM{Example scan \#3.}{}}
\end{subfigure}
\begin{subfigure}{0.1875\textwidth}\centering
\stackunder{\includegraphics[width=\linewidth]{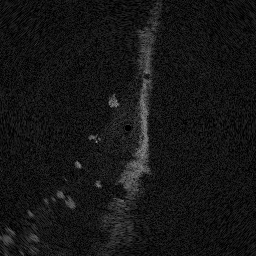}}{\fixM{Example scan \#4.}{4.2.4}}
\end{subfigure}

\caption{\label{fig:twolochs_overview}
The \texttt{twolochs} dataset is the most extensive we provide, and is collected over a journey featuring (1) a tarmac road under tree cover alongside a \textit{Loch Laggan} (top), (2) a gravel track over the shoulder of a mountain, (3) a partially submerged track fording a loch, (4) a gravel track alongside \textit{Lochan na h-Earba}, and (5) a gravel track descending a mountainside to the start location.
}
\end{subfigure}

\caption{
\fixM{The \texttt{maree} and \texttt{twolochs} datasets, with GPS traces, ground truth matrices, and sample scans.}{4.1}
}
\label{fig:dataset_overview_b}
\end{figure*}

\begin{table}
\centering
\resizebox{\columnwidth}{!}{
\begin{tabular}{c|cccc}
Dataset & Variable terrain & Poor weather & Scene diversity & Repeat traversals\\
\hline
Radar RobotCar & \ding{55} & \ding{55} & \ding{55} & \ding{51} \\
Marulan & \ding{51} & \ding{51} & \ding{51} & \ding{51} \\
MulRan & \ding{55} & \ding{55} & \ding{51} & \ding{51} \\
RADIATE & \ding{55} & \ding{51} & \ding{51} & \ding{55} \\
Boreas & \ding{55} & \ding{51} & \ding{51} & \ding{51} \\
OSDaR23 & \ding{55} & \ding{55} & \ding{51} & \ding{55} \\
\textit{OORD (Ours)} & \ding{51} & \ding{51} & \ding{51} & \ding{51}
\end{tabular}
}
\caption{
Comparison of features of our dataset in comparison to other radar datasets in literature.
\textit{Marulan} is closest in characteristics to ours, but is collected with a short-range sensor, \SI{40}{\metre} as opposed to our \SI{165}{\metre} range.
}
\label{tab:vs_other_datasets}
\end{table}

\section{Related Work}
\label{sec:related}

\subsection{Radar datasets}

Datasets featuring a similar class of millimetre-wave scanning radar include \textit{Marulan}~\cite{peynot2010marulan} by Peynot \textit{et al}, which is furthermore also collected in natural environments, as is ours.
This data, however, only has range of up to \SI{40}{\metre}.
Several recent radar datasets use the same radar class -- from the same manufacturer\footnote{\textit{Navtech Radar}: \rurl{navtechradar.com}} -- as used in our work, including the \textit{Oxford Radar RobotCar Dataset}~\cite{barnes2020oxford} by Barnes \textit{et al}, \textit{MulRan}~\cite{kim2020mulran} by Kim \textit{et al}, \textit{RADIATE}~\cite{sheeny2021radiate} by Sheeny \textit{et al}, \textit{Boreas}~\cite{burnett2023boreas} by Burnett \textit{et al}, and \textit{OSDaR23}~\cite{tagiew2023osdar23} by Tagiew \textit{et al}.
The \textit{Oxford Radar RobotCar Dataset}~\cite{barnes2020oxford} is purely urban and features many repeat traversals, useful in investigating place recognition.
The ground truth provided in~\cite{barnes2020oxford}, however, is prepared for radar odometry rather than place recognition.
\textit{MulRan}~\cite{kim2020mulran} presents more diversity in scenery, having been collected in several distinct parts of Daejeon, South Korea.
We similarly attend to generalisation-sensitive  training/testing data and collect at many sites, albeit offroad and more widespread across a large area of countryside/wilderness rather than a city.
Our repeat traversals are also more numerous than for \textit{MulRan}.
\textit{RADIATE}~\cite{sheeny2021radiate}, presenting vehicle detection and tracking ground truth, is similarly lacking in overlapping traversals and not geared towards place recognition, although a diversity in weather makes this dataset interesting for exploring this sensor's robustness.
Closest to our work is \textit{Boreas} by Burnett \textit{et al}~\cite{burnett2023boreas}, who collect multi-season data including a \textit{Navtech CIR304-H} scanning radar.
Our work is in part inspired by this, with ours featuring highly non-planar offroad driving as opposed to purely urban. 

\cref{tab:vs_other_datasets} summarises the characteristics of our dataset as compared to these.

\subsection{Radar place recognition}

Owing to its inherent resilience in sensing capabilities, radar has attracted considerable attention for its adeptness in the face of adverse weather, demonstrating a capacity to navigate and operate effectively in inclement conditions.
This is because radar systems typically use electromagnetic waves in the radio or microwave frequency range.
Unlike visible light, these waves can penetrate through various weather elements like rain, snow, fog, and clouds without significant distortion or attenuation.
Radar place recognition has thus been explored 
in~\cite{suaftescu2020kidnapped,barnes2020under,gadd2020look,wang2021radarloc,komorowski2021large,gadd2021contrastive,hong2022radarslam,adolfsson2023tbv,jang2023iros,yuan2023iros}.

This has included simultaneous localisation and mapping~\cite{hong2022radarslam,adolfsson2023tbv} with loop closure detection mechanisms, where for example in~\cite{hong2022radarslam} loop candidates are rejected by principal components analysis to focus on distinct structural layouts.

There are also several methods for learning to vectorise radar scans with neural networks
~\cite{suaftescu2020kidnapped,barnes2020under,gadd2021contrastive,wang2021radarloc,komorowski2021large,yuan2023iros}, including supervised~\cite{suaftescu2020kidnapped,barnes2020under,wang2021radarloc,wang2021radarloc} and self-supervised approaches~\cite{yuan2023iros,gadd2021contrastive} and learned uncertainty~\cite{yuan2023iros}. 

Finally, some approaches are non-learned
\cite{kim2020mulran,gadd2020look,jang2023iros,gadd2024open}, instead relying 
largely on transformation-vinvariant integral transforms (e.g. Fourier, Fourier-Mellin, Radon) as in~\cite{jang2023iros,gadd2024open}, pooling operations such as max/min/mean to achieve rotation invariance and to reduce the representation dimensionality~\cite{kim2020mulran,gadd2024open}, and circular cross-correlation for an alignment/similarity score~\cite{kim2020mulran,jang2023iros}.

\fixM{
This place recognition capability can be crucial for simultaneous localisation and mapping (SLAM), where \textit{reinitialisation} is essential for correcting errors and maintaining accuracy over time.
SLAM is a newly developed area for scanning radar in autonomous vehicle operations, as per~\cite{hong2020radarslam,wang2022maroam,adolfsson2023tbv}, and so robust reinitialisation through radar place recognition is an important research topic.
}{3.1.1}
This area is therefore of increasing interest, but far from mature, and this dataset should therefore contribute to establishing useful benchmarks for the community.

\section{Off-road routes \& collection sites}
\label{sec:routes}

\cref{fig:dataset_overview_a,fig:dataset_overview_b} show a detailed overview of the datasets we provide, at various areas of the \textit{Ardverikie Estate}.
Data has been gathered in the region encompassing \textit{Loch Laggan} and \textit{Lochan na h-Earba}. 
The position of these lochs can be particularly well understood by referring to the caption of~\cref{fig:twolochs_overview}.

\textit{Loch Laggan}, a freshwater loch located about \SI{10}{\kilo\metre} west of \textit{Dalwhinnie} in the Scottish Highlands, extends in a nearly northeast to southwest direction for approximately \SI{11}{\kilo\metre}.
The name \textit{Lochan na h-Earba} actually refers to two lochs positioned south of \textit{Loch Laggan}.
These lochs are situated in a slender glen that runs from southwest to northeast, running roughly parallel to \textit{Loch Laggan}.
The \textit{Binnein Shuas} range of hills separates them from \textit{Loch Laggan}.

Each of~\cref{fig:twolochs_overview,fig:bellmouth_overview,fig:maree_overview,fig:hydro_overview} is arranged as follows: \textit{Top Left}: Photograph of each dataset collection location (handheld camera, not on-board imagery).
\textit{Top Middle}: GPS trace of route driven.
\textit{Top Right}: Ground truth match matrix for a pair of trajectories at that site.
\textit{Bottom}: Example radar scans from various points along the route (Cartesian).

\begin{figure}
\centering
\includegraphics[width=\columnwidth]{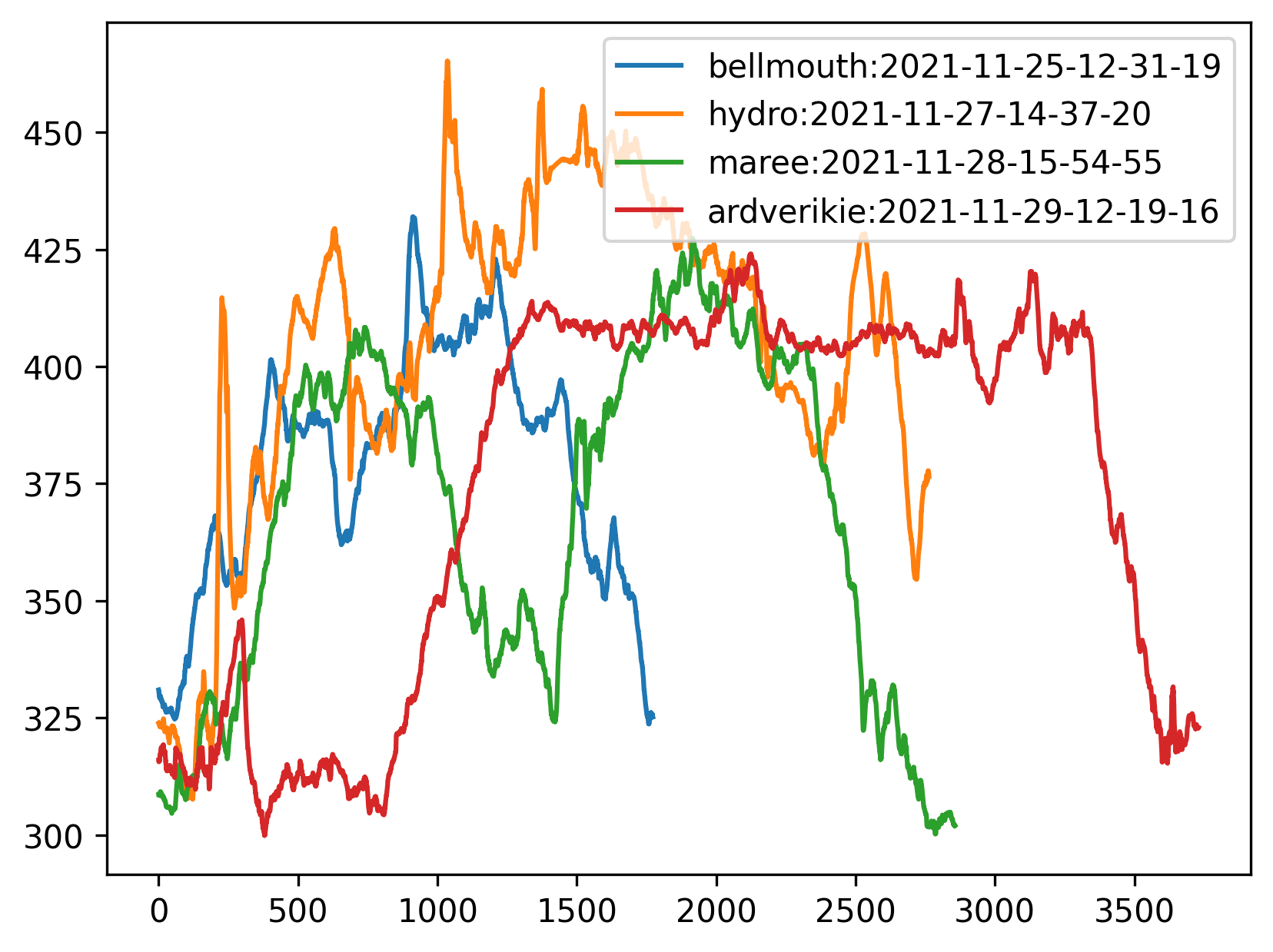}
\caption{
Changes in elevation (measured as UTM down) for example forays from each challenge site.
\fixM{The horizontal axis shows time elapsed while driving (\SI{}{\second}) and the vertical axis shows UTM down (\SI{}{\metre}) as a measure of elevation.}{4.3}
}
\label{fig:elevation}
\end{figure}

We release data from four areas of the \textit{Ardverikie Estate}.
These feature distinct landscape (therefore typical radar returns) as well as driving conditions.
We refer to the sites as \textit{Two Lochs}, \textit{Bellmouth}, \textit{Maree}, and \textit{Hydro}.
Each of these is discussed below.
\cref{tab:summary_table} summarises the collection sites and routes.
\cref{fig:elevation} compares the elevation along each route.

\begin{table}[!h]
\centering
\resizebox{\columnwidth}{!}{
\begin{tabular}{c|ccccc}
Route &  Distance & \# Radar & \fixM{\# Camera}{3.2.5} & \# GPS & Size \\
\hline
Bellmouth $\times4$ & \SI{8.98}{\kilo\metre} & $7781$ & \fixM{$6647$}{} & $7091$ & \SI{6.65}{\gibi\byte} \\
Hydro $\times3$ & \SI{13.65}{\kilo\metre} & $11094$ & \fixM{$13832$}{} & $11054$ & \SI{8.93}{\gibi\byte}  \\
Maree $\times2$ & \SI{16.87}{\kilo\metre} & $11462$ & \fixM{$10724$}{} & $11427$ & \SI{9.05}{\gibi\byte}  \\
Two Lochs $\times2$ &  \SI{20.62}{\kilo\metre} & $14938$ & \fixM{$11239$}{} & $14922$ & \SI{11.70}{\gibi\byte} 
\end{tabular}
}
\caption{
\fixM{Summary statistics for the various routes to the sites they represent.
\textbf{Quantities in each row are from an example foray on that route} -- \textit{the total distance driven in our dataset is approximately \SI{154}{\kilo\metre} and approximately \SI{90}{\gibi\byte}.} 
Note that for ``Camera'' the framerate has been downsampled to more approximately match the radar frame rate (and mitigate download sizes).
}{4.1.2}
}
\label{tab:summary_table}
\end{table}

\subsection{Route 1: Bellmouth}
\label{sec:bellmouth}

This is a \SI{8.98}{\kilo\metre} route.
The route is completely landlocked, heading away from and then back towards \textit{Loch Laggan}, see~\cref{fig:bellmouth_overview} (\textit{Top Middle}.
The ascent and descent are reflected by a somewhat symmetrical elevation change in~\cref{fig:elevation} -- climbing for the first half of the route, and descending for the second half. 
The vehicle is always on a gravel track.
\cref{bfly:2021-11-25-12-31-19,bfly:2021-11-25-12-01-20,bfly:2021-11-26-16-12-01,bfly:2021-11-26-15-35-34} show images from the vehicle's perspective from four outings along this route.
The first two these outings is prior to any snow falling in the \textit{Ardverikie Estate} area (\cref{bfly:2021-11-25-12-01-20,bfly:2021-11-25-12-31-19}).
The next two of these outings show significant snow dump (\cref{bfly:2021-11-26-15-35-34,bfly:2021-11-26-16-12-01}).
The outings are in the same direction, as per the ground truth matrix in~\cref{fig:twolochs_overview} (\textit{Top Right}), but within each outing itself we double back (c.f. the off diagonal matches in three parts of this matrix).

\subsection{Route 2: Hydro}
\label{sec:hydro}

This is a \SI{13.65}{\kilo\metre} route.
As shown in~\cref{fig:elevation} this route reaches the highest elevation of all or our collection sites -- ascending a gravel track to maximum elevation and then descending along the same route.
For parts of the route, we travel along a mountain river, as per~\cref{fig:hydro_overview} (\textit{Top Middle}).
The snow coverage is significant (\cref{bfly:2021-11-27-14-37-20,bfly:2021-11-27-15-24-02,bfly:2021-11-27-16-03-26}), but not as thick as for \textit{Maree} below.

\subsection{Route 3: Maree}
\label{sec:maree}

This is a \SI{16.87}{\kilo\metre} route.
The route is mountainous, in the elevated area to the south of \textit{Loch Laggan}, see~\cref{fig:maree_overview} (\textit{Top Middle}).
There are two ascents and descents as can be seen in~\cref{fig:elevation} -- with a descent halfway through the journey, and another at the end of the journey. 
The vehicle is primarily on a gravel track.
\cref{bfly:2021-11-28-15-54-55,bfly:2021-11-28-16-43-37} show images from the vehicle's perspective from two outings along this route.
Both of these feature heavy snow, with a completely covered track.
In \cref{bfly:2021-11-28-16-43-37} the vehicle is driving in total darkness (with no artificial illumination beyond the vehicle's spotlights as this is a very remote and isolated natural collection site).
The outings are in the same direction, as per the ground truth matrix in~\cref{fig:maree_overview} (\textit{Top Right}), but within each outing itself we double back on parts of the route and return to the starting point along the same route we took to begin the route (c.f. the off diagonal matches in three parts of this matrix).

\subsection{Route 4: Two Lochs}
\label{sec:twolochs}

This is a \SI{20.62}{\kilo\metre} route across the entire estate.
The route is alongside two lochs (\textit{Loch Laggan} and \textit{Lochan na h-Earba}), see~\cref{fig:twolochs_overview} (\textit{Top Middle}, where the lochs are on top of and the bottom of the image, respectively).
Referring to the route in an anti-clockwise sense, while going alongside \textit{Loch Laggan} the vehicle is on tarmac -- it is otherwise on unpaved road for the rest of the route.
\cref{bfly:2021-11-29-12-19-16,bfly:2021-11-29-11-40-37} show images from the vehicle's perspective of our two outings -- these are completed \textit{after} snow has fallen (in previous days), so melted somewhat, but while it is still very visible on the ground etc.
The outings are in the opposite direction as shown by the off-diagonal matches in the ground truth matrix in~\cref{fig:twolochs_overview} (\textit{Top Right}).
Note that the red elevation profile in~\cref{fig:elevation} corresponds to the foray shown in~\cref{fig:twolochs_overview} (\textit{Top Middle}), and is very flat at approximately \SI{400}{\metre} as the vehicle traverses alongside \textit{Lochan na h-Earba}.

\subsection{Summary: Datasets provided}
\label{sec:routes_summary}

In summary, the datasets provided are listed below, including the date-string for the collection time and the site that they were collected at, with colours corresponding to~\cref{fig:elevation}.

\begin{itemize}
\item  \fixM{}{4.4.1}{\crtcrossreflabel{\textcolor{black}{(\fixC{Bellmouth \#1})}}[DB1]}: \texttt{2021-11-25-12-01-20} \fixS{(Bellmouth)}{}
\item  {\crtcrossreflabel{\textcolor{black}{(\fixC{Bellmouth \#2})}}[DB2]}: \texttt{2021-11-25-12-31-19} \fixS{(Bellmouth)}{}
\item  {\crtcrossreflabel{\textcolor{black}{(\fixC{Bellmouth \#3})}}[DB3]}: \texttt{2021-11-26-15-35-34} \fixS{(Bellmouth)}{}
\item  {\crtcrossreflabel{\textcolor{black}{(\fixC{Bellmouth \#4})}}[DB4]}: \texttt{2021-11-26-16-12-01} \fixS{(Bellmouth)}{}
\item  {\crtcrossreflabel{\textcolor{black}{(\fixC{Hydro \#1})}}[DH1]}: \texttt{2021-11-27-14-37-20} \fixS{(Hydro)}{}
\item  {\crtcrossreflabel{\textcolor{black}{(\fixC{Hydro \#2})}}[DH2]}: \texttt{2021-11-27-15-24-02} \fixS{(Hydro)}{}
\item  {\crtcrossreflabel{\textcolor{black}{(\fixC{Hydro \#3})}}[DH3]}: \texttt{2021-11-27-16-03-26} \fixS{(Hydro)}{}
\item  {\crtcrossreflabel{\textcolor{black}{(\fixC{Maree \#1})}}[DM1]}: \texttt{2021-11-28-15-54-55} \fixS{(Maree)}{}
\item  {\crtcrossreflabel{\textcolor{black}{(\fixC{Maree \#2})}}[DM2]}: \texttt{2021-11-28-16-43-37} \fixS{(Maree)}{}
\item  {\crtcrossreflabel{\textcolor{black}{(\fixC{Two Lochs \#1})}}[DA1]}: \texttt{2021-11-29-11-40-37} \fixS{(Two Lochs)}{}
\item {\crtcrossreflabel{\textcolor{black}{(\fixC{Two Lochs \#2})}}[DA2]}: \texttt{2021-11-29-12-19-16} \fixS{(Two Lochs)}{}
\end{itemize}

\begin{figure}[!h]
\centering
\begin{subfigure}{0.32\columnwidth}
\includegraphics[width=\textwidth]{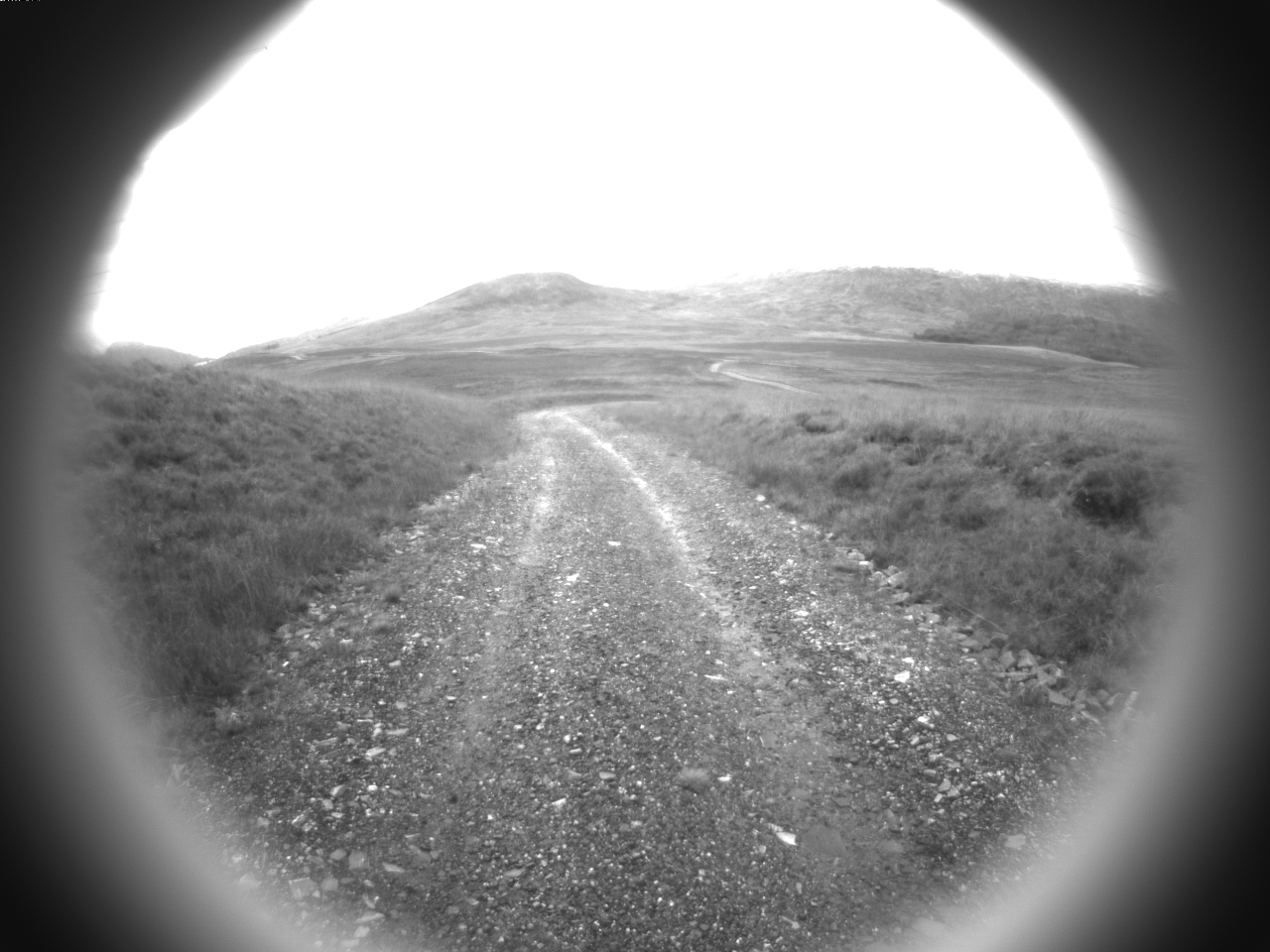}
\caption{\fixM{}{4.4.3}\ref{DB1}\label{bfly:2021-11-25-12-01-20}}
\end{subfigure}
\begin{subfigure}{0.32\columnwidth}
\includegraphics[width=\textwidth]{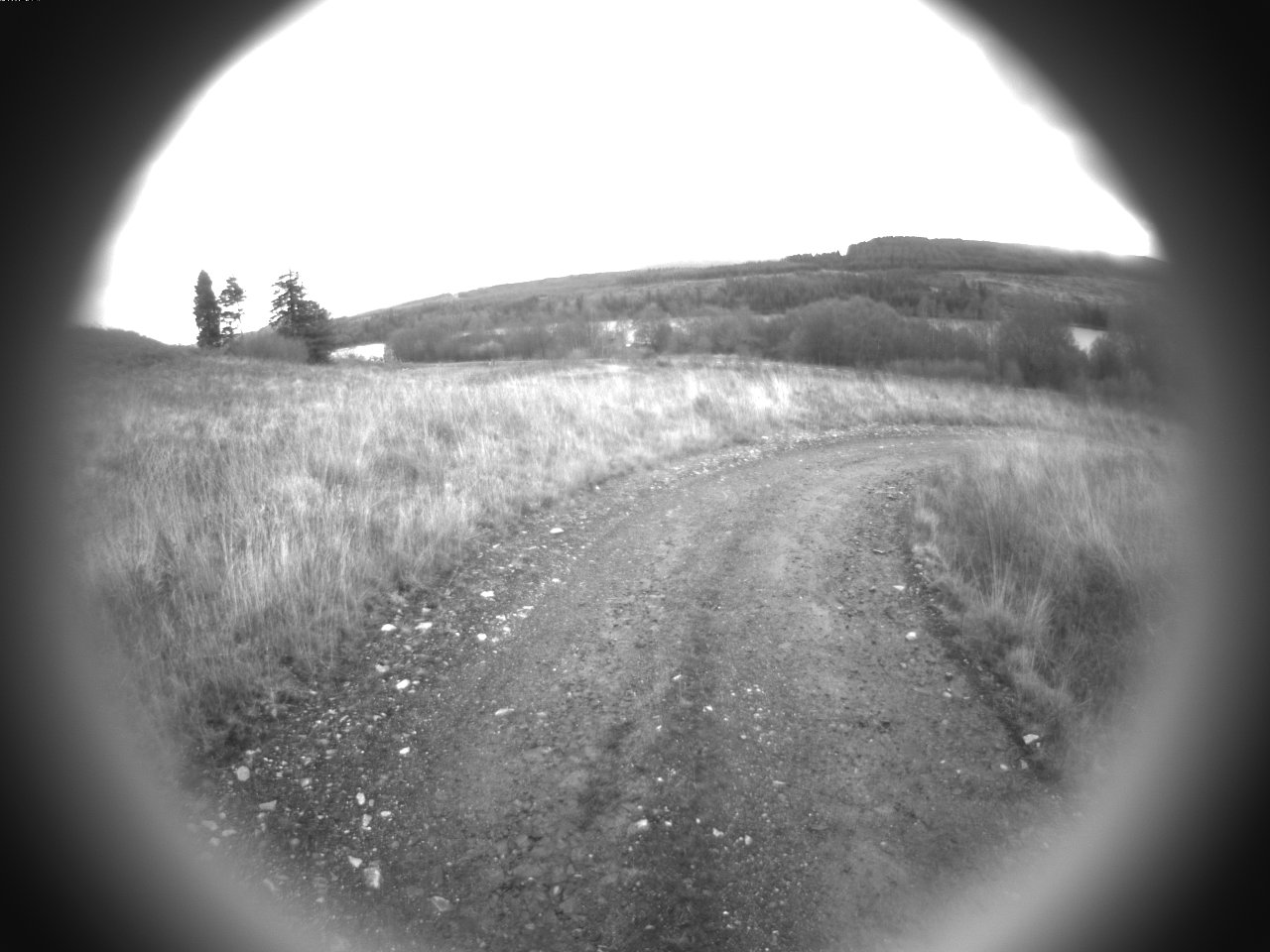}
\caption{\ref{DB2}\label{bfly:2021-11-25-12-31-19}}
\end{subfigure}
\begin{subfigure}{0.32\columnwidth}
\includegraphics[width=\textwidth]{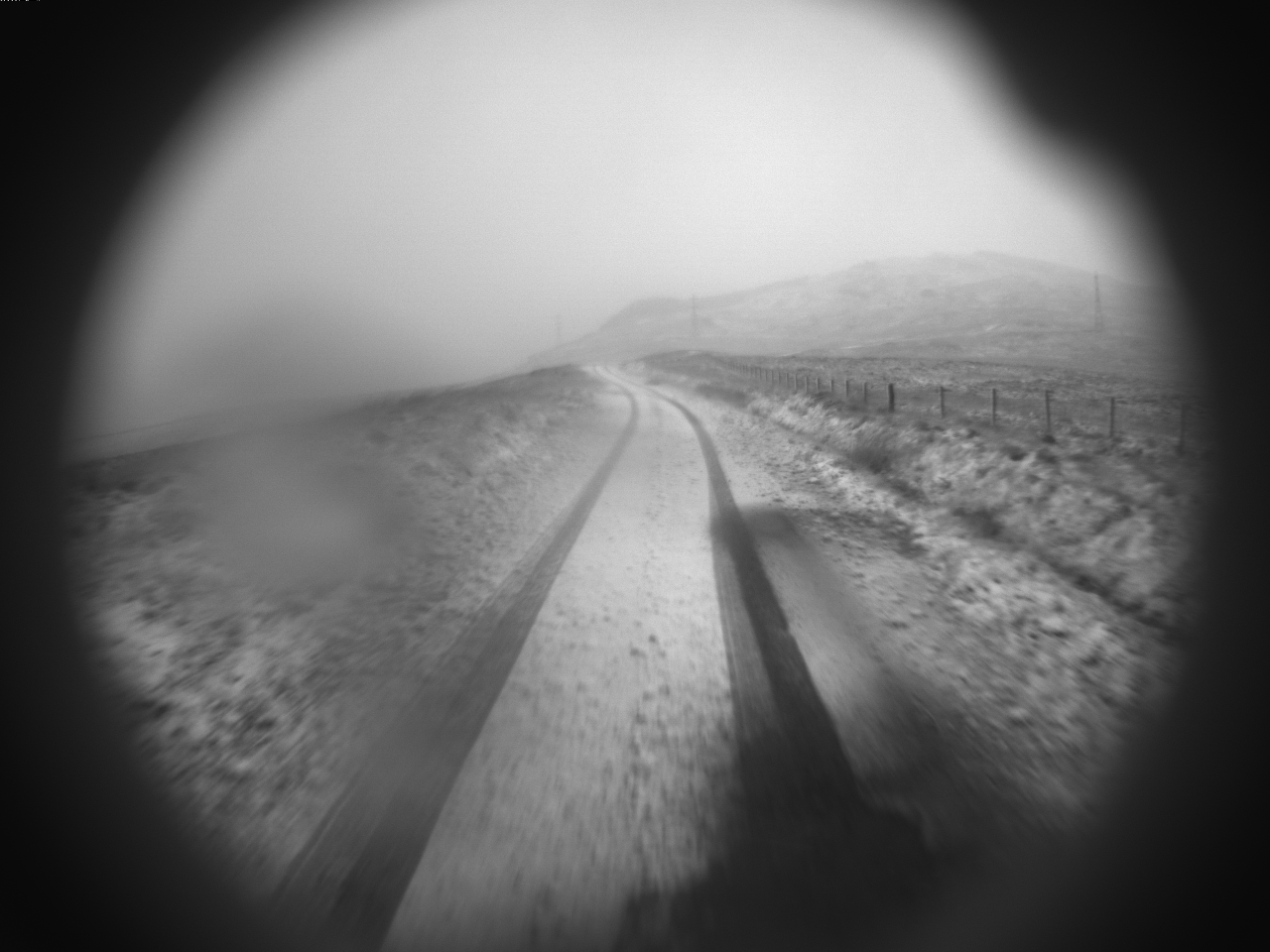}
\caption{\ref{DB3}\label{bfly:2021-11-26-15-35-34}}
\end{subfigure}

\begin{subfigure}{0.32\columnwidth}
\includegraphics[width=\textwidth]{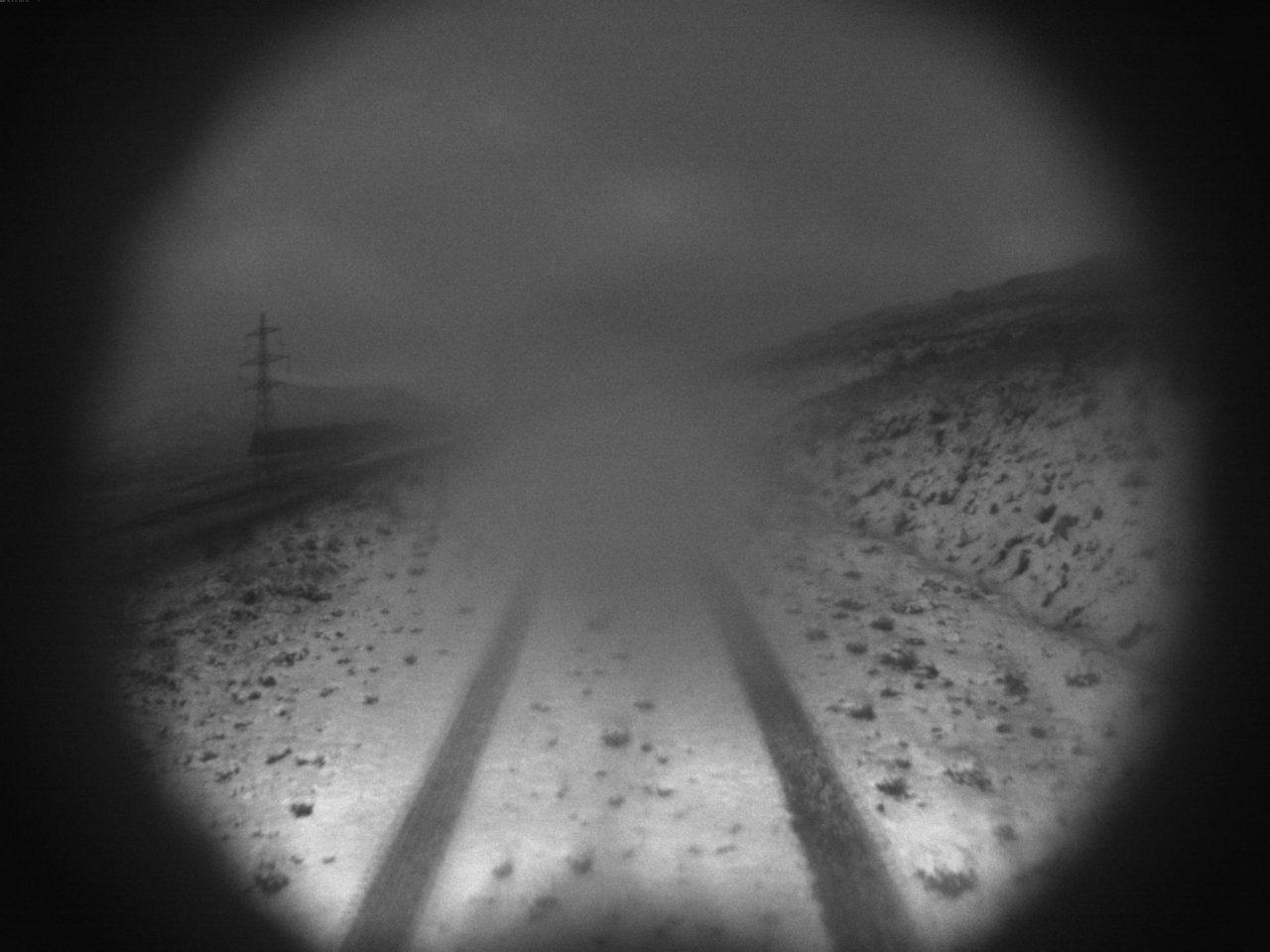}
\caption{\ref{DB4}\label{bfly:2021-11-26-16-12-01}}
\end{subfigure}
\begin{subfigure}{0.32\columnwidth}
\includegraphics[width=\textwidth]{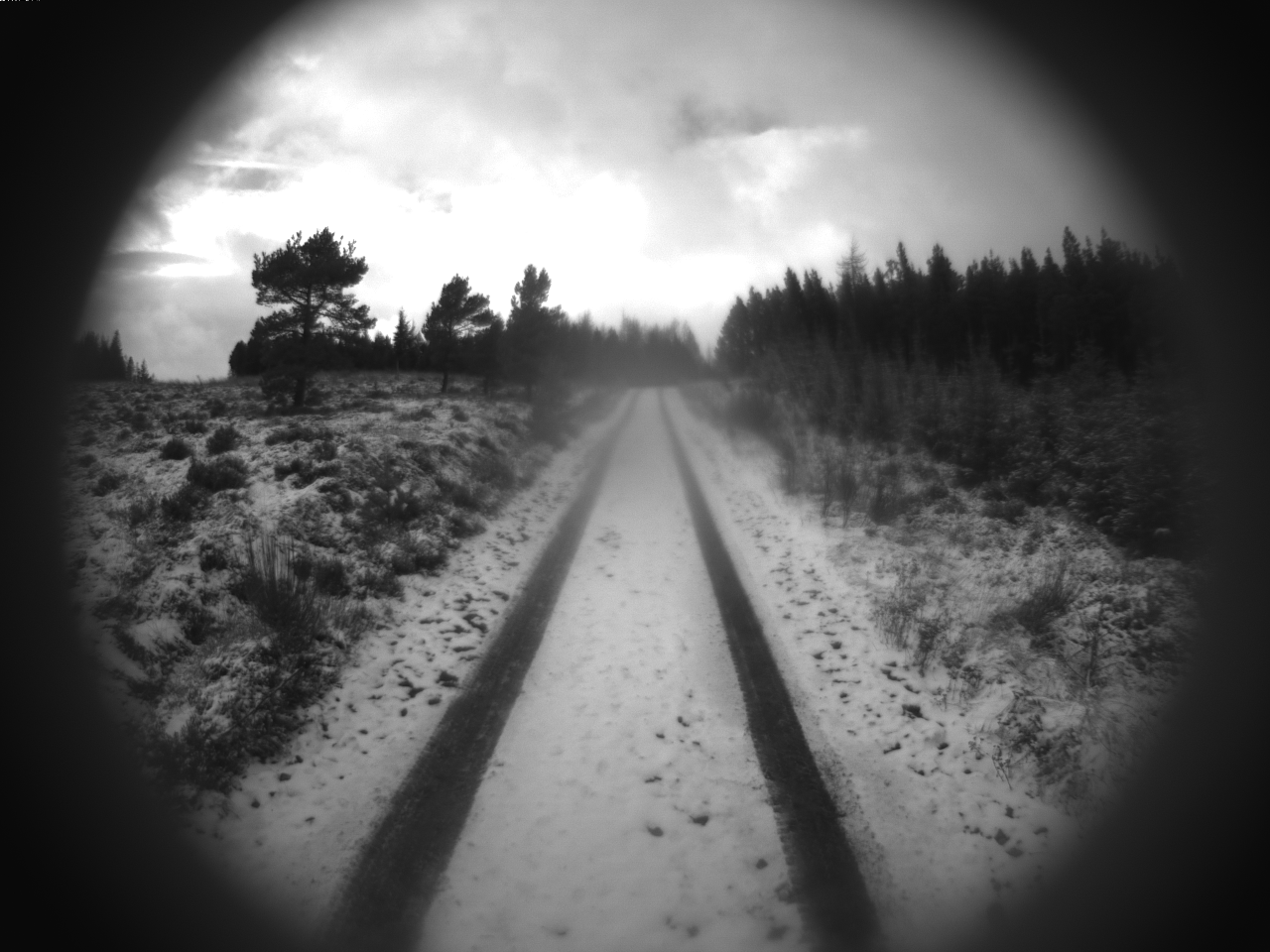}
\caption{\ref{DH1}\label{bfly:2021-11-27-14-37-20}}
\end{subfigure}
\begin{subfigure}{0.32\columnwidth}
\includegraphics[width=\textwidth]{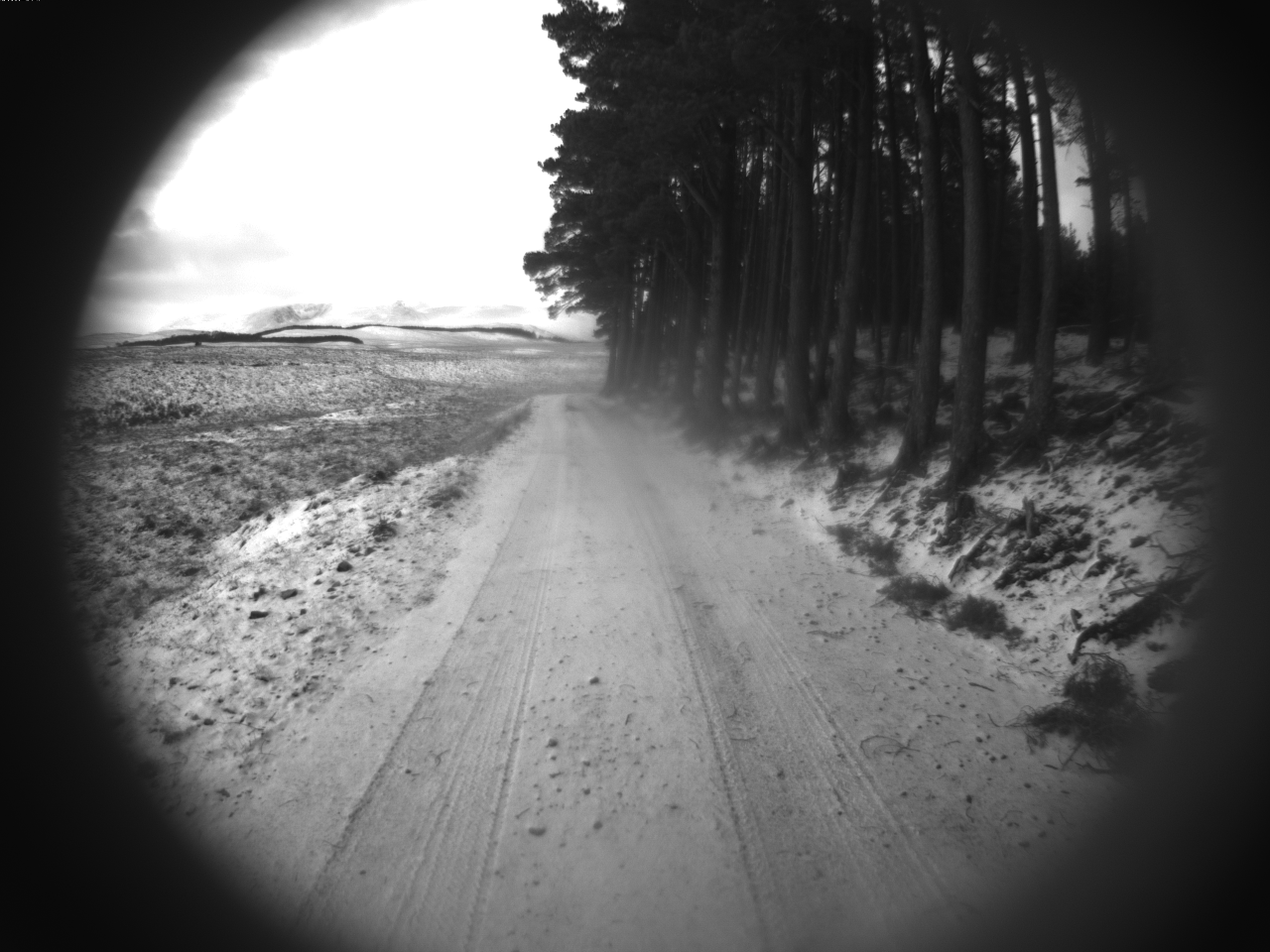}
\caption{\ref{DH2}\label{bfly:2021-11-27-15-24-02}}
\end{subfigure}

\begin{subfigure}{0.32\columnwidth}
\includegraphics[width=\textwidth]{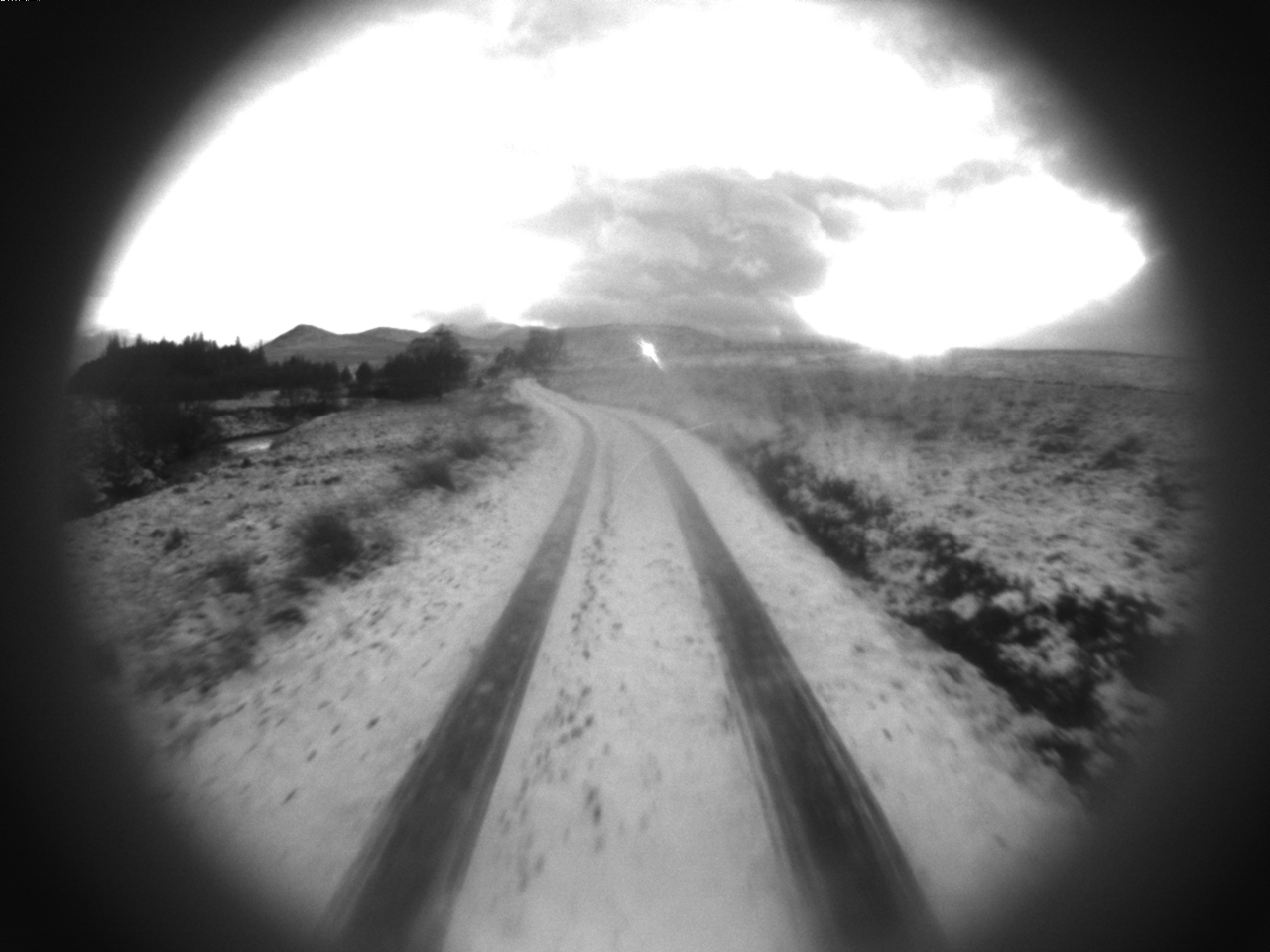}
\caption{\ref{DH3}\label{bfly:2021-11-27-16-03-26}}
\end{subfigure}
\begin{subfigure}{0.32\columnwidth}
\includegraphics[width=\textwidth]{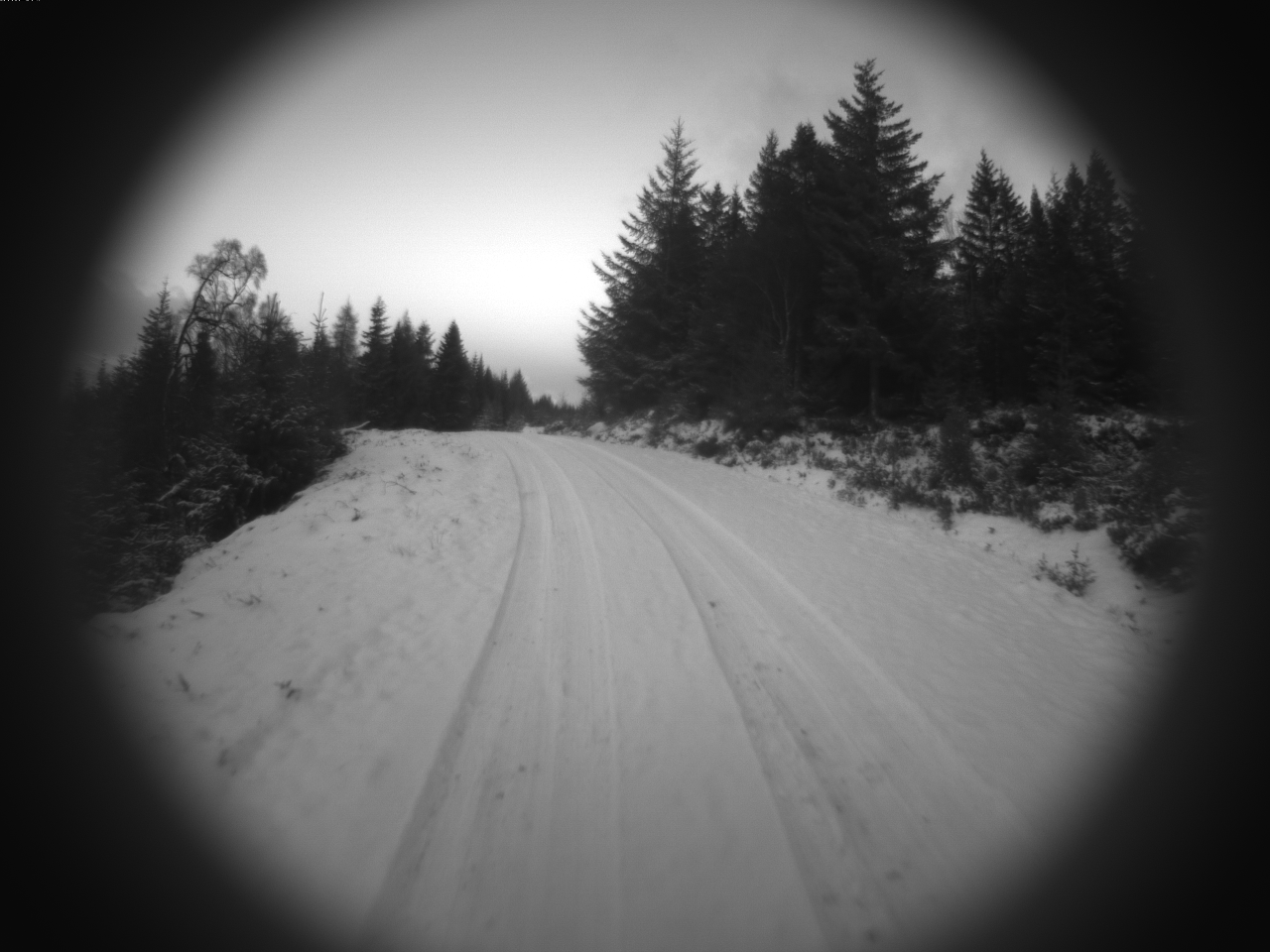}
\caption{\ref{DM1}\label{bfly:2021-11-28-15-54-55}}
\end{subfigure}
\begin{subfigure}{0.32\columnwidth}
\includegraphics[width=\textwidth]{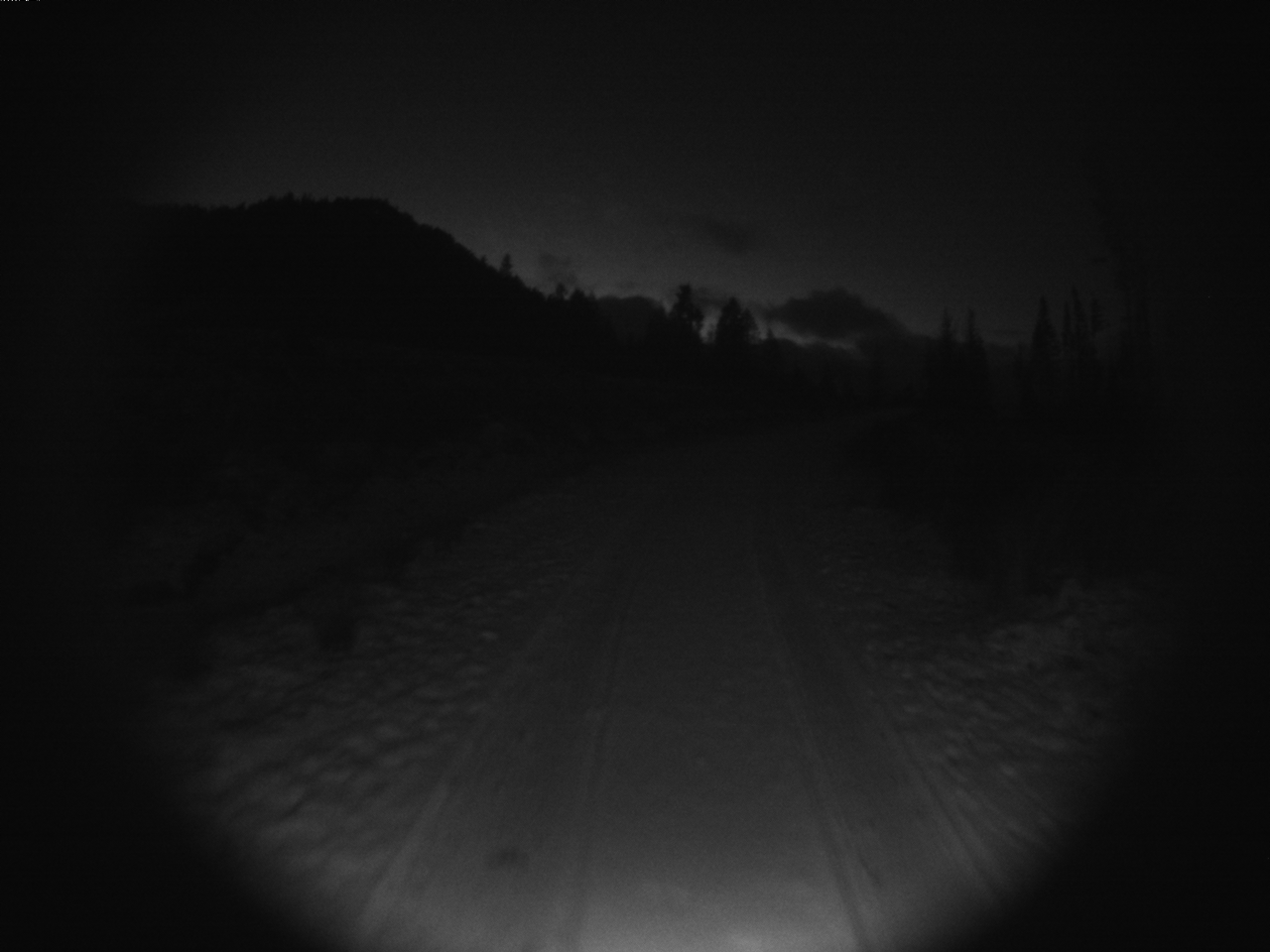}
\caption{\ref{DM2}\label{bfly:2021-11-28-16-43-37}}
\end{subfigure}

\begin{subfigure}{0.32\columnwidth}
\includegraphics[width=\textwidth]{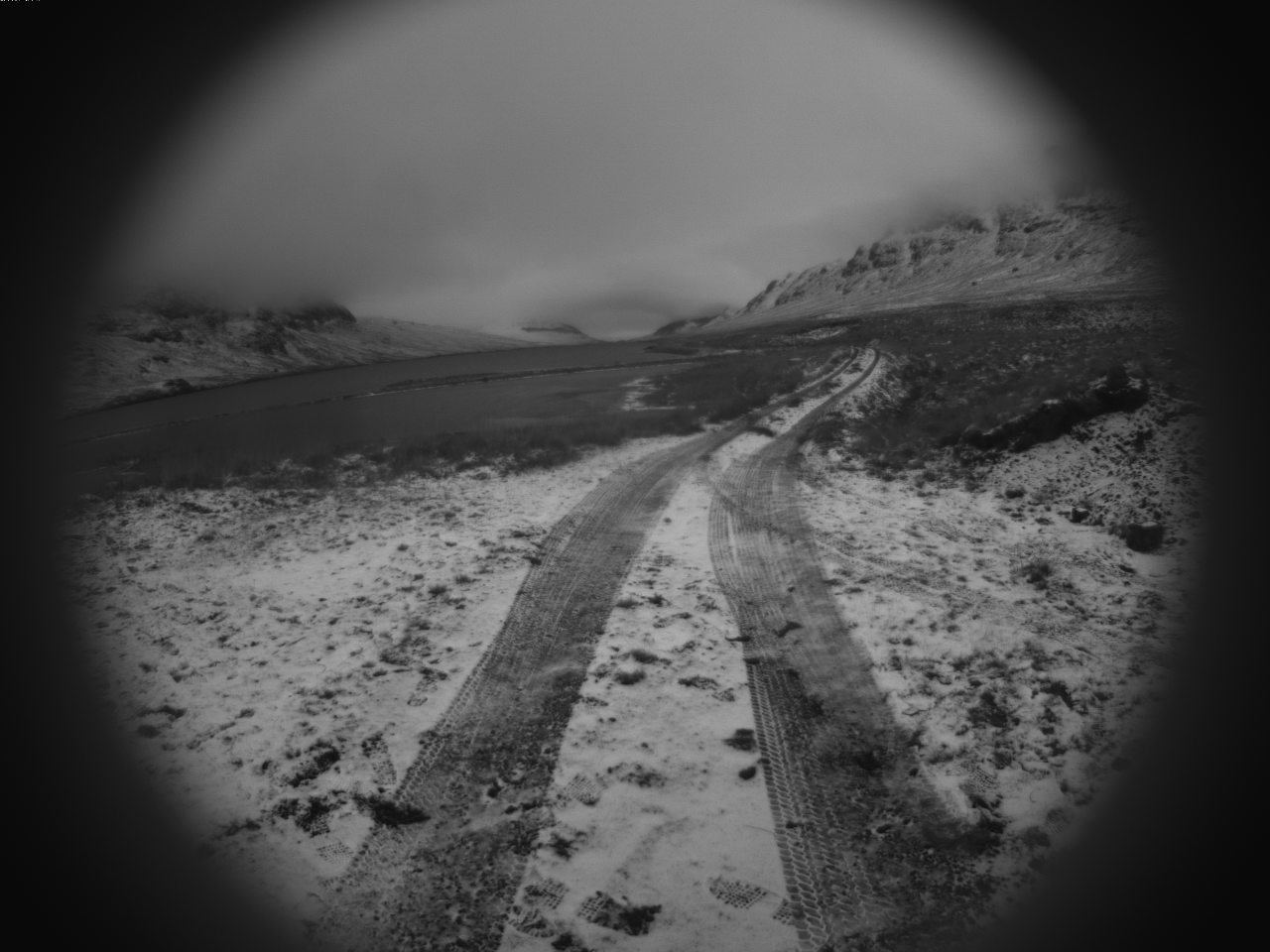}
\caption{\ref{DA1}\label{bfly:2021-11-29-11-40-37}}
\end{subfigure}
\begin{subfigure}{0.32\columnwidth}
\includegraphics[width=\textwidth]{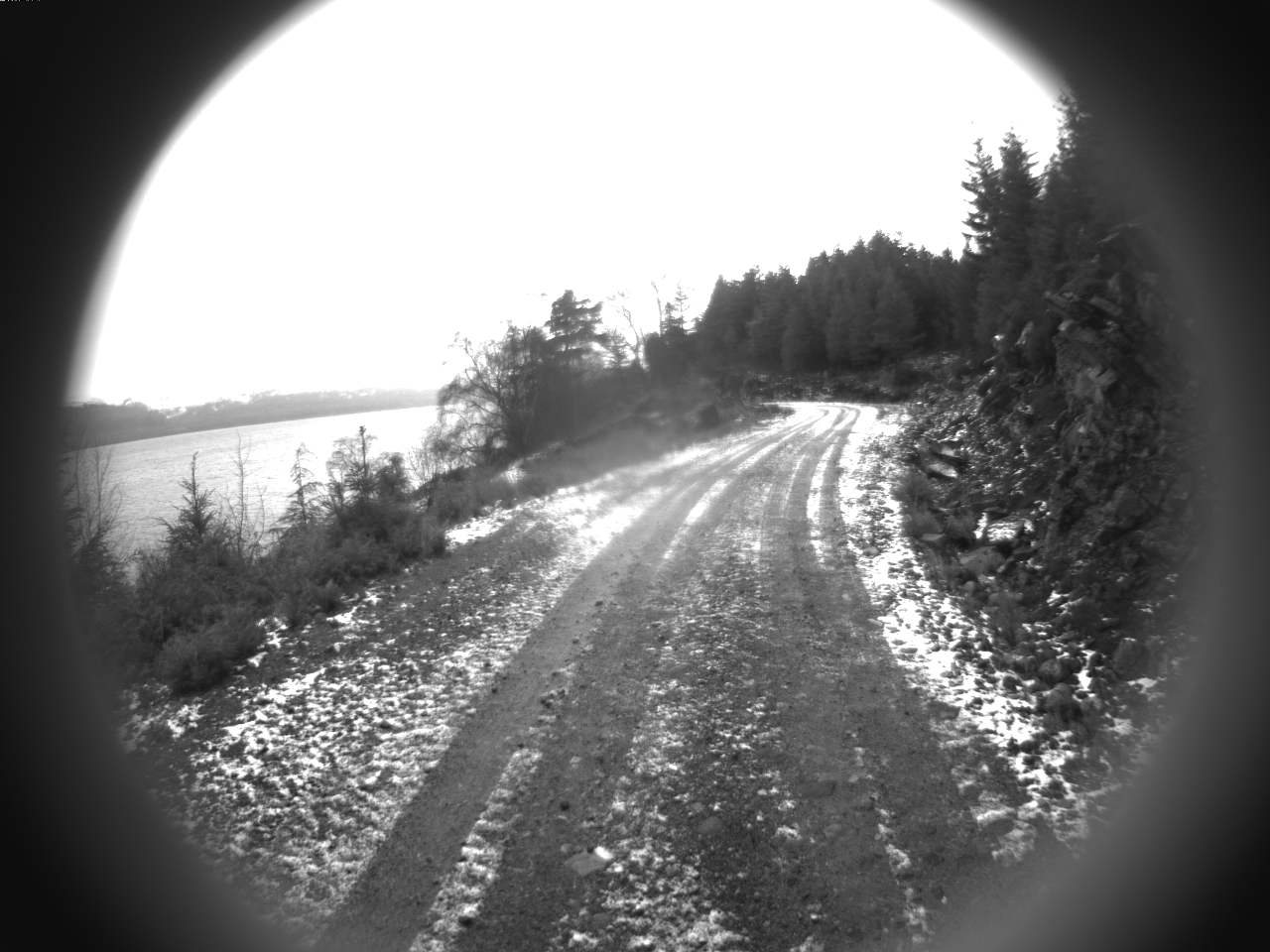}
\caption{\ref{DA2}\label{bfly:2021-11-29-12-19-16}}
\end{subfigure}

\caption{
\fixM{On-board camera images for each released foray in~\cref{sec:routes_summary}.
Note that as this dataset is radar-focused, we do not release these camera images, but they are included here to give a sense of the inclement collection conditions.
}{4.1.3}
\label{fig:BFLY}
}
\end{figure}

These are shown in~\cref{fig:BFLY}.
We note that, beyond the weather and illumination effects discussed for each route above, some forays have lens adherents on the on-board camera (e.g. \cref{bfly:2021-11-26-15-35-34,bfly:2021-11-26-16-12-01} in particular), motivating the use of radar in this sort of domain and for the localisation task.

\section{Sensors \& dataset format}
\label{sec:dataset}

At \rurl{oxford-robotics-institute.github.io/oord-dataset} we provide download time-stamped synchronised radar and GPS/INS.

\subsection{Radar}
\label{sec:dataset:radar}

This sensor is the same as that used for the \textit{Oxford Radar RobotCar Dataset}~\cite{barnes2020oxford}, i.e. a Navtech CTS350-X Millimetre-Wave frequency-modulated continuous wave radar, \SI{4}{\hertz}, \SI{400}{} measurements per rotation, \SI{163}{\metre} range, \SI{4.38}{\centi\metre} range resolution, \SI{1.8}{\degree}~beamwidth.
The radar was mounted at the centre of the vehicle aligned to the vehicle axes.

Radar scans are stored as lossless-compressed PNG files in polar form with each row representing the sensor reading at each azimuth and each column representing the raw power return at a particular range.

\fixM{\subsection{Camera}
\label{sec:dataset:bfly}
\fixM{}{3.2.3}
Our vehicle was also equipped with a \textit{FLIR Blackfly USB3 Vision} camera\footnote{\rurl{flir.com/products/blackfly-s-usb3}} which captures images at $1920\times1200$ resolution over a $11.25\times7.03$~\SI{}{\milli\metre} sensing area and with \SI{67.71}{\deci\bel} of dynamic range.

Some example images are shown in~\cref{fig:BFLY}, and we present experimental results comparing vision to radar in~\cref{sec:results}.
Note that in~\cref{tab:summary_table} the camera framerate has been downsampled to more approximately match the radar frame rate (and mitigate download sizes).

This will be useful for sensor fusion research such as~\cite{wang2021rodnet}, where our radar and camera are timestamp-synchronised.
}{}

\subsection{GPS/INS}
\label{sec:dataset:gps}

We use a Microstrain 3DM-RQ1-45 GPS/INS.
This sensor has direct satellite and inertial measurements, and computes position, velocity, and attitude.
To provide accurate readings, it uses a triaxial accelerometer, gyroscope, magnetometer, and temperature sensors, as well as a pressure altimeter.
Furthermore, dual on-board processors run a Extended Kalman Filter (EKF).

\fixM{
\subsection{Calibration}
\label{sec:dataset:calibration}
\fixM{}{3.3}
Extrinsic calibration between the sensors is done by reference to a detailed CAD model of the vehicle and sensor suite.
These are included in the SDK (\cref{sec:tools}) as an \texttt{extrinsics.yaml} file specifying the relative six degree-of-freedom pose between each of the GPS/INS (\cref{sec:dataset:gps})
\begin{equation}
\mathbf{t}_{\mathrm{radar},\mathrm{gps}}=(-0.7634,0,0,0,0,0)
\end{equation}
which is to say that the GPS unit is approximately \SI{763}{\milli\metre} behind the radar with axes aligned to the radar.
Then, for the camera (\cref{sec:dataset:bfly}) to radar
\begin{equation}
\mathbf{t}_{\mathrm{radar},\mathrm{camera}}=(-1.494,0.302,0,0,-\frac{\pi}{2},\frac{\pi}{2})
\end{equation}
which is to say that the camera is further behind and to the right of the radar with axes offset by two \SI{90}{\degree} rotations.
Finally, the radar (\cref{sec:dataset:radar}, as the main sensor focused on in this work) is at $(0,0,0,0,0,0)$, i.e. is used as the datum/origin of the frame of reference.
}{}

\section{Software \& experimental tools}
\label{sec:tools}

At \rurl{github.com/mttgdd/oord-dataset} and \rurl{huggingface.co/mttgdd/oord-models} we provide a software tools to support the use of this dataset.

\subsection{Downloads \& data loaders}
\label{sec:tools:data}

Similarly to~\cite{barnes2020oxford}, we provide data loaders to work with the radar data format, which as per~\cref{sec:dataset:radar} are released in raw polar format, but can be easily transformed into Cartesian format with our SDK.

For working with driven sequences, we provide a basic iterable \texttt{Dataset} class which returns each radar scan paired with its \textit{closest} (in a timestamp sense) GPS/INS reading (\cref{sec:dataset:gps}).

We also provide a batch download script to rapidly access the data.

\subsection{Radar place recognition \& evaluation}
\label{sec:tools:evaluation}

\fixM{
In place recognition it is common to represent the input image or scan by some descriptor vector.
With map descriptors $\mathbf{m}^j\in\mathbb{R}^{D}$ at places indexed by $j$ and of dimension $D$ and a query descriptor $\mathbf{q}\in\mathbb{R}^{D}$, place recognition can be stated as
\begin{equation}\label{eq:nn}
j^*=\operatorname{argmin}_j||\mathbf{q}-\mathbf{m}_j||_2^2
\end{equation}
where we decide that we are in place $j^*$ given that the descriptor for that place is closest to our query descriptor (i.e. ``place recognition'').
Alternatively, instead of~\cref{eq:nn}, we may search for a set $N$ of nearest neighbours.
}{3.1.2}

We evaluate place recognition performance by comparing ``distance matrices'' in the ground truth space given by GPS, as shown by example in~\cref{fig:dataset_overview_a,fig:dataset_overview_b} (\textit{Top Right}) to distance matrices composed of e.g. pairwise distances between vectors representing radar scans (see~\cref{sec:examples}) or some other similarity measurement.
These matrices have heights given by the number of radar scans in a ``query'' foray and widths similarly for a ``reference'' foray.

\fixM{
Specifically, with respect to~\cref{eq:nn} above we have black and white distance matrices as inin~\cref{fig:dataset_overview_a,fig:dataset_overview_b} (\textit{Top Right}) given by
\begin{equation}
\mathtt{GT}_{ij}=||\mathbf{g}_i-\mathbf{g}_j||_2
\label{eqn:gtm}
\end{equation}
where $\mathbf{g}\in\mathbb{R}^2$ is the UTM northing/easting GPS coordinates corresponding to the query/localisation trajectory scan $i$ and reference/map trajectory scan $j$, with $\mathtt{GT}\in\mathbb{R}^{RL}$ and with $R$ and $L$ as the number of frames in these trajectories, respectively.
Similarly, we have
\begin{equation}
\mathtt{DM}_{ij}=||\mathbf{q}_i-\mathbf{m}_j||_2
\label{eqn:dm}
\end{equation}
with $\mathbf{q}$ and $\mathbf{m}$ embeddings of the query/reference radar scans at those same locations $i,j$.
After reducing $\mathtt{DM}$ to its binarised form $\mathtt{DM}_{\mathtt{0/1}}$ with $1$ in each row indicating the nearest neighbour (or, generally, a set of $1$s indicate some $N$ nearest neighbours) as per~\cref{eq:nn}, we expect for good place recognition that $\mathtt{DM}_{\mathtt{0/1}}\approx\mathtt{GT}$ (with a numerical accuracy defined as \texttt{Recall@N} below).
}{1.3.1}

Highlighted regions are such that those radar scan pairs have locations given by GPS which are within \SI{25}{\metre} (this threshold being configurable in our SDK).
In these, interesting features to note are the off-diagonal revisits, which represent traversals in the opposite direction.
This is especially important in the radar modality as the sensor has a \SI{360}{\degree} field-of-view.

\begin{figure}
\centering

\begin{subfigure}{0.48\columnwidth}
\includegraphics[width=\textwidth]{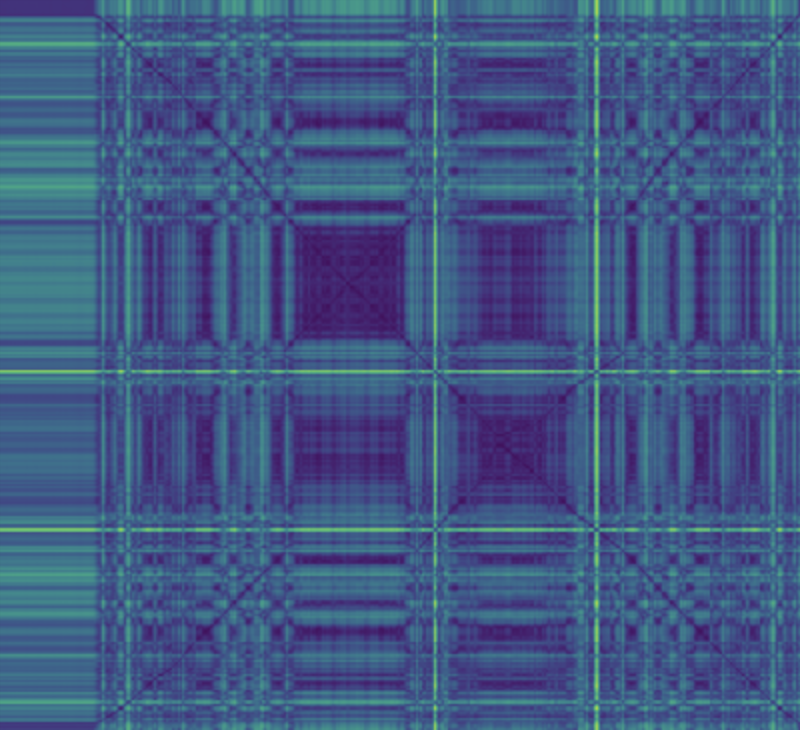}
\caption{\textbf{RingKey}, \ref{B1}\label{figs:diff_ringkey}}
\end{subfigure}
\begin{subfigure}{0.48\columnwidth}
\includegraphics[width=\textwidth]{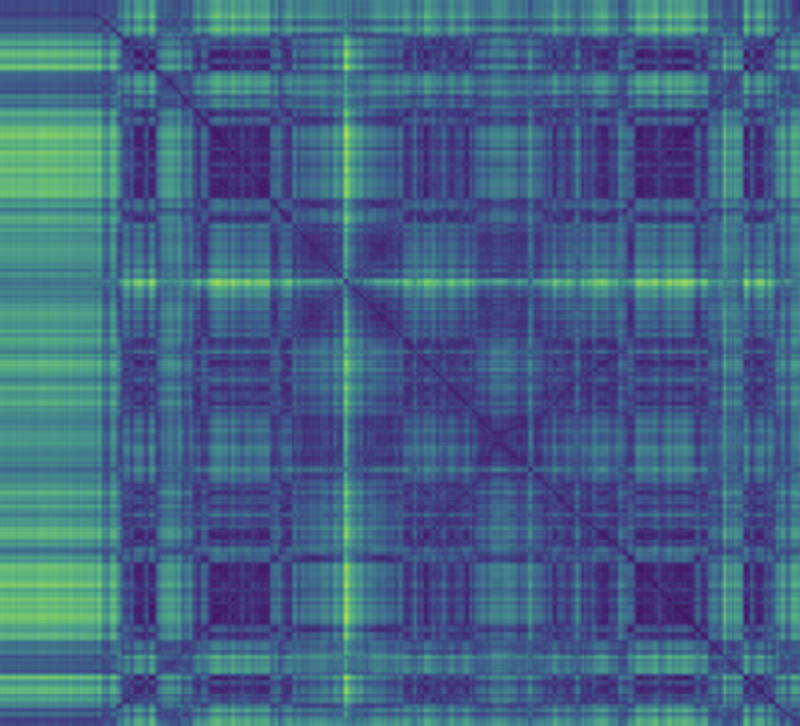}
\caption{\textbf{AlexNet}, \ref{B8}\label{figs:diff_alexnet}}
\end{subfigure}

\begin{subfigure}{0.48\columnwidth}
\includegraphics[width=\textwidth]{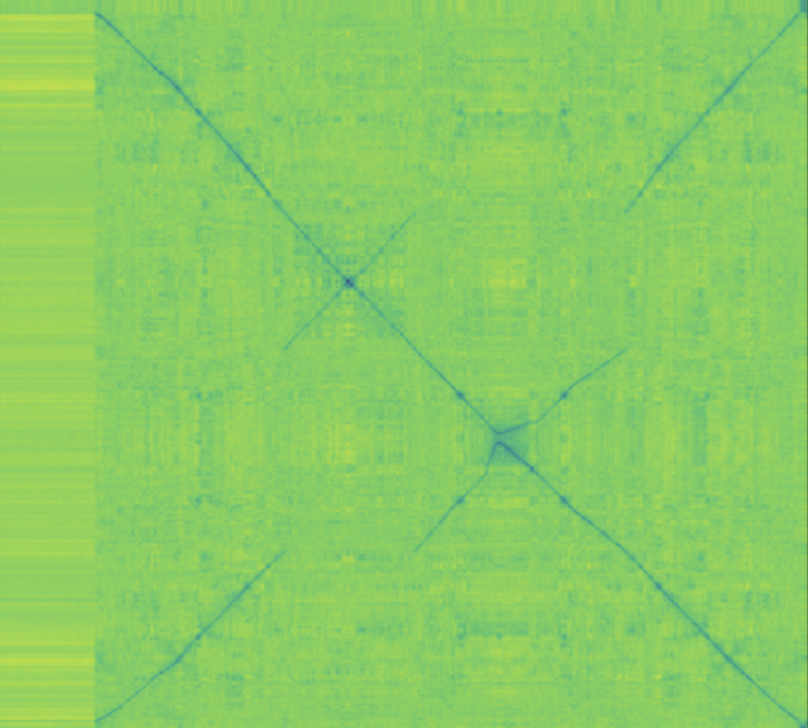}
\caption{\textbf{RaPlace}, \ref{B2}\label{figs:diff_raplace}}
\end{subfigure}
\begin{subfigure}{0.48\columnwidth}
\includegraphics[width=\textwidth]{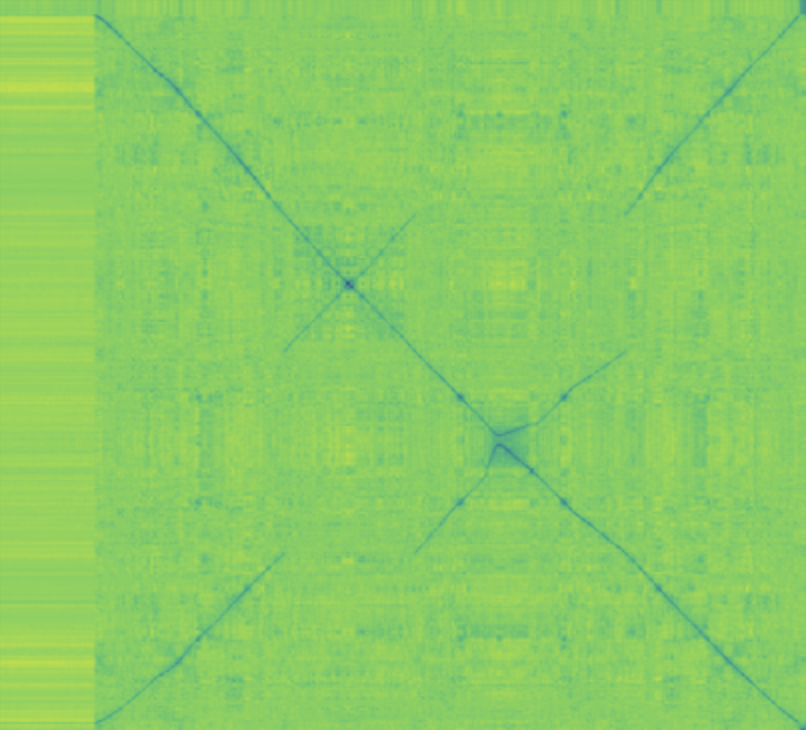}
\caption{\textbf{RadVLAD}, \ref{B3}\label{figs:diff_radvlad}}
\end{subfigure}

\caption{
Example embedding distance matrices between a live trajectory (rows) and a reference trajectory (columns) in this case \texttt{2021-11-25-12-31-19} and \texttt{2021-11-25-12-01-20} from \texttt{bellmouth}, respectively.
\ref{B2} and \ref{B3} are the most distinctive, being radar-specific methods.
\fixM{
Here, \subref{figs:diff_ringkey} and \subref{figs:diff_alexnet} are ``noisier'' than \subref{figs:diff_raplace} and \subref{figs:diff_radvlad} and not as good an approximation of the \texttt{bellmouth} ground truth difference matrix in \cref{fig:bellmouth_overview} (\textit{Top Right}) because \ref{B1} and \ref{B2} produce less discriminative/informative embeddings of radar scans for the radar place recognition task.
}{1.3.2}
\label{fig:diff}
}
\end{figure}

Then, \texttt{Recall@N} refers to the percentage of query scans which result in \textit{at least one} successful localisation (i.e. resulting in a returned location within \SI{25}{\metre} of the query location) when accepting as candidates all of the $N$ nearest neighbours in the embedding space.
This is a commonly used metric and threshold, as in visual place recognition~\cite{arandjelovic2016netvlad,berton2022rethinking}.

\fixM{
Consider the black and white matrices shown in
\cref{fig:twolochs_overview,fig:bellmouth_overview,fig:maree_overview,fig:hydro_overview} (\textit{Top Right}).
Each of these has, as per~\cref{eqn:gtm} and~\cref{eqn:dm}, a number of rows equal to the number of scans (with some examples in the bottom row of \cref{fig:dataset_overview_a,fig:dataset_overview_b}) collected along a query route (shown by the red GPS traces in \cref{fig:dataset_overview_a,fig:dataset_overview_b}) and number of columns equal to the number of scans collected along a reference route -- i.e. the reference route acts as a map and the query route is localised against it.
Each cell in these matrices are thus a pair of scans from a pair of locations.
The black background means that the distance between those locations is $>$\SI{25}{\metre} and white entries indicate that the pair of locations is $<$\SI{25}{\metre}.
}{1.2}
Note that in~\cref{fig:dataset_overview_a,fig:dataset_overview_b} the true positives are ``thickened'' for visualisation purposes, to \SI{100}{\metre}, but we experiment in~\cref{sec:results} with \SI{25}{\metre}.

For this we use a ``difference matrix'' in~\cref{fig:diff} with the embedding distances (Euclidean) between live and map features.
\fixM{Not that the ``grids'' apparent in \cref{figs:diff_ringkey,figs:diff_alexnet} are artefacts resulting from embeddings representing scans being very similar to many such embeddings in the map.
These are evident as ``blocks'' of blue in the difference matrix.
Within this block, all embeddings are very similar (where yellow indicates embeddings being more dissimilar).
This phenomena is therefore known as ``perceptual ambiguity'' or ``aliasing''.
This indicates poor representational capability of the method, where we want only scans from the same location to be similar -- see~\cref{figs:diff_raplace,figs:diff_radvlad}, in which only the diagonal and lines perpendicular to the diginal are blue, with no such gridlike structure.
}{1.3.3}
Localisation is successful if the nearest neighbour for a query embedding (rows) is a reference embedding (column) which is in truth (c.f. the ground truth matrices in~\cref{fig:bellmouth_overview} \textit{Top Right}) is close to the query location in physical space.

\subsection{Radar place recognition with neural networks}
\label{sec:tools:nns}

\fixM{
As introduced in~\cref{sec:tools:evaluation}, radar place recognition is concerned with learning discriminative and compressed representations of radar scans, to maximise similarity between likewise places and minimise similarity otherwise.
However, for this and similar sensors -- despite their inherent immunity to appearance change -- the beam is not narrow,
power returns are perturbed by noises (e.g. speckle) and artefacts (multi-path reflections, etc).
However, radar data -- despite being complex in this way -- contains highly informative information.
While it is difficult to manually/visually inspect radar data for this information, it has been found in~\cite{suaftescu2020kidnapped,gadd2020look,yuan2023iros,gadd2021contrastive} that neural networks are able to learn to encode very informative feature representations that are germane to the task of place recognition. 
}{2.1}
As will be discussed below, we evaluate a family of learned methods on our dataset.
This includes a pretrained and frozen backbone feature extractor along with a pooling layer which is fine-tuned on radar data from this dataset, and we release these weights at \rurl{huggingface.co/mttgdd/oord-models}\kern-.5ex, with this being, to our knowledge, the first open-sourced learned radar place recognition system.


\subsection{Open-source radar place recognition methods}
\label{sec:tools:benchmarks}

As below in~\cref{sec:examples}, our SDK includes either (1) new implementations of or (2) imports, as submodules, of other open-source radar place recognition systems~\cite{kim2020mulran,jang2023raplace,gadd2024open}.
Experiments on novel systems can be quickly run over all pairs of trajectories discussed in~\cref{sec:routes_summary}, and prepared in the format of~\cref{tab:examples}, by a common \texttt{.yaml} configuration format.

\section{Examples}
\label{sec:examples}

We demonstrate the utility of our dataset as a novel radar place recognition benchmark using the evaluation method described in~\cref{sec:tools:evaluation}.
We also show that the GPS/INS described in~\cref{sec:dataset:gps} can be put to good use in \textit{supervising} learned radar place recognition methods.
Therefore, we demonstrate implementations of the following methods with our dataset \fixM{, including four methods, \ref{B1}--\ref{B4}, which are either not fine-tuned at all (by virtue of the method itself) or are fine-tuned on this data as well as four methods which are totally pre-trained, \ref{B5}--\ref{B9}, i.e. not fine-tuned on this new dataset.}{1.4.5}

\fixM{\textit{Non-learned radar-specific methods}}{4.5.1}

\begin{itemize}
\item {\crtcrossreflabel{\textcolor{black}{(Baseline 1)}}[B1]}:
{\bf RingKey} as a component of \textit{ScanContext}~\cite{kim2018scan} averages point cloud contents at a fixed distance around the vehicle (thus being rotation invariant).
Note, for this we do not use the orientation refinement of \textit{ScanContext}, as distances between row-reduced vectors and orientation scores are separate, i.e. these form a hierarchy of localisers.
To be clear, this is different to~\ref{B2} below, where the maximum circular cross-correlation is the score used for similarity between scans (with no initial score used to reduce the candidate set).
\item {\crtcrossreflabel{\textcolor{black}{(Baseline 2)}}[B2]}:
{\bf RaPlace}~\cite{jang2023raplace} measures the similarity score between radar scans by using Radon-transformed sinogram images and cross-correlation in the frequency domain.
This gives rigid transform invariance during place recognition, and supresses the effects of radar multipath and ring noises.
\item {\crtcrossreflabel{\textcolor{black}{(Baseline 3)}}[B3]}:
{\bf Open-RadVLAD}~\cite{gadd2024open} uses only the polar representation.
Also, for partial translation invariance and robustness to signal noise, it uses only a 1D Fourier Transform along radial returns.
It also achieves rotational invariance and a very discriminative descriptor space by building a vector of locally aggregated descriptors (VLAD).
\end{itemize}

\fixM{\textit{Fine-tuned radar-specific methods}}{4.5.2}

\begin{itemize}
\item {\crtcrossreflabel{\textcolor{black}{(Baseline 4)}}[B4]}:
Here we train radar-specific neural network models -- using \textbf{ResNet18}~\cite{he2016deep} as a backbone feature extractor and \textbf{NetVLAD}~\cite{arandjelovic2016netvlad} as a pooling layer (with $64$ clusters).
Inputs are polar radar scans of $128\times128$ resolution and embeddings are $128$-dimensional.
The radar returns are replicated $3$ times at the input in order to form a $3$-channel image. 
We start with pretrained weights learned against \textbf{ImageNet}~\cite{deng2009imagenet} and then, in an unsupervised learning step, initialise the \textbf{NetVLAD} cluster centres by k-means++~\cite{arthur2007k} over the deep features extracted from the reference trajectory radar frames.
\end{itemize}

\fixM{\textit{Pre-trained methods}}{4.5.3}

\begin{itemize}
\item {\crtcrossreflabel{\textcolor{black}{(Baseline 5)}}[B5]}, {\crtcrossreflabel{\textcolor{black}{(Baseline 6)}}[B6]}, {\crtcrossreflabel{\textcolor{black}{(Baseline 7)}}[B7]}, {\crtcrossreflabel{\textcolor{black}{(Baseline 8)}}[B8]}, {\crtcrossreflabel{\textcolor{black}{(Baseline 9)}}[B9]}:
Here we show that the plug-and-play ability of our dataset with respect to widely available pretrained networks, to accelerate research in this area.
We use specifically the \texttt{pytorch/vision:v0.10.0} models available at \rurl{pytorch.org/hub} as well as the \texttt{gmberton/cosplace} models available at \rurl{github.com/gmberton/CosPlace}.
This covers \textbf{MobileNet} \cite{sandler2018mobilenetv2} \ref{B6}, \textbf{GoogleNet}~\cite{szegedy2015going} \ref{B7}, \textbf{AlexNet}~\cite{krizhevsky2012imagenet} \ref{B8}, and \textbf{VGG19}~\cite{simonyan2014very} \ref{B9}, which are all trained for \textit{ImageNet} classification and thus not specialised for place recognition but represent inputs in terms of wide variety of concepts and features.
It also covers \textbf{CosPlace (ResNet18-512)}~\cite{berton2022rethinking} \ref{B5} which is specifically trained for place recognition -- but not specifically for the radar modality, in contrast to \ref{B4}.
\end{itemize}

\fixM{
The pretrained networks are selected as they have been applied to \textit{visual} place recognition in other work -- e.g. \textbf{MobileNet} by Wu \textit{et al.}~\cite{wu2019deep}, \textbf{AlexNet} by S{\"u}nderhauf \textit{et al.}~\cite{sunderhauf2015performance}, \textbf{GoogleNet} by Yue \textit{et al.}~\cite{yue2015exploiting}, \textbf{VGG} by Arandjelovic \textit{et al.}~\cite{arandjelovic2016netvlad}, and \textbf{ResNet} by Yu \textit{et al.}~\cite{yu2019spatial} -- and it
is therefore important to apply them to place recognition in the radar domain.
}{1.4.1}
\noindent\cref{sec:results} presents a comprehensive comparison of these example baselines over all pairs of trajectories from our dataset.
\fixM{
A comparison of the computational efficiency of these various baselines is outside of the scope of this work, but the interested reader is referred to~\cite{canziani2016analysis} (Figure 2, page 2) for an in-depth comparison of inference speeds and memory requirements for the pretrained networks \ref{B5}--\ref{B9} and to~\cite{gadd2024open} for a comparison of the radar-specific methods, \ref{B1}--\ref{B3}, where \textbf{Open-RadVLAD} is shown to be more computationally efficient than \textbf{RaPlace}.
}{1.4.7}

\begin{table*}
\centering
\resizebox{\textwidth}{!}{
\renewcommand{\arraystretch}{1.2}
\begin{tabular}{l|c|c|c|c|c}
\fixM{}{4.4.5} & \ref{DB2}-vs-\ref{DB1} & \ref{DH1}-vs-\ref{DH2} & \ref{DH1}-vs-\ref{DH3} & \ref{DM1}-vs-\ref{DM2} & \ref{DA2}-vs-\ref{DA1}\\
\hline
\ref{B1} & 94.29 / 97.53 / 98.24 & 91.15 / 93.57 / 93.78 & 94.06 / 95.80 / 96.00 & 91.76 / 97.22 / 98.42 & 72.96 / 82.28 / 84.71\\
\ref{B2} & 98.45 / 98.73 / 98.94 & 92.65 / 93.89 / 93.94 & 95.34 / 95.85 / 95.85 & 94.28 / 99.15 / 99.40 & \textbf{85.82} / \textbf{87.69} / \textbf{88.25}\\
\ref{B3} & \textbf{98.87} / \textbf{99.79} / \textbf{99.93} & 93.51 / 94.10 / 94.16 & \textbf{96.00} / 96.16 / 96.16 & 96.88 / \textbf{99.06} / 99.32 & 84.27 / 87.36 / \textbf{88.25}\\
\ref{B4} & 97.67 / 98.52 / 98.87 & \textbf{93.78} / \textbf{94.26} / \textbf{94.37} & 95.70 / \textbf{99.59} / \textbf{99.95} & \textbf{97.14} / \textbf{99.06} / \textbf{99.44} & 5.41 / 10.98 / 16.23\\
\ref{B5} & 90.49 / 97.25 / 98.10 & 92.76 / 93.83 / 93.83 & 94.72 / 95.85 / 95.85 & 89.11 / 97.91 / 98.72 & 4.53 / 10.10 / 15.51\\
\ref{B6} & 90.70 / 97.04 / 98.45 & 91.96 / 93.62 / 93.83 & 93.14 / 95.49 / 95.75 & 87.92 / 96.67 / 98.25 & 17.55 / 40.07 / 50.44\\
\ref{B7} & 94.50 / 98.38 / 98.80 & 92.28 / 93.83 / 93.83 & 93.70 / 95.54 / 95.80 & 90.61 / 97.27 / 98.42 & 23.18 / 47.19 / 58.22\\
\ref{B8} & 74.91 / 92.39 / 95.28 & 82.79 / 91.31 / 92.60 & 83.91 / 92.78 / 94.62 & 65.16 / 86.93 / 92.10 & 14.74 / 35.87 / 47.08\\
\ref{B9} & 72.73 / 90.49 / 94.71 & 81.29 / 90.40 / 91.96 & 84.07 / 93.14 / 94.31 & 69.43 / 87.15 / 91.97 & 9.71 / 21.80 / 31.29\\\end{tabular}
}
\caption{
\texttt{Recall@1/5/10} localisation success rate (\%) for all example methods defined in~\cref{sec:examples} and discussed in~\cref{sec:results} over $5$ pairs of trajectories from our dataset.
}
\label{tab:examples}
\end{table*}

\section{Results}
\label{sec:results}

\begin{figure}
\centering
\includegraphics[width=\columnwidth]{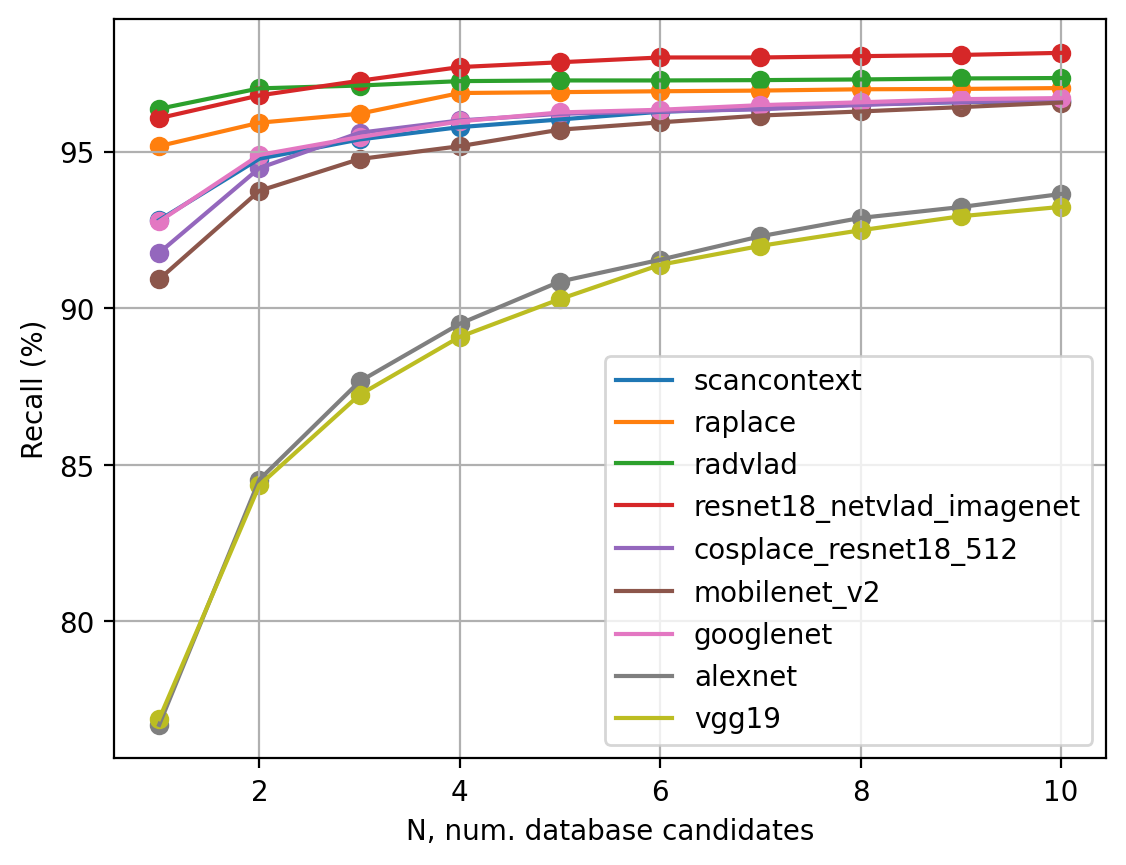}
\caption{
\texttt{Recall@N} localisation success rate, as described in~\cref{sec:tools:evaluation}.
\ref{B2} (\textbf{RaPlace}), \ref{B3} (\textbf{RadVLAD}), and \ref{B4} (\textbf{ResNet18-NetVLAD}) perform best.
The approach customised to radar data, \ref{B4}, outstrips others when more database candidates are retrieved. 
}
\label{fig:rc}
\end{figure}

\fixM{\subsection{Baseline comparison}}{}
\fixM{}{4.5.4}
As can be seen in~\cref{tab:examples}, 
considering that all of each of the routes is at minimum \SI{9}{\kilo\metre} in length (see\cref{tab:summary_table}), all radar-specific methods perform very well, with in excess of \SI{90}{\percent} for even \texttt{Recall@1} i.e. when relying on only a single nearest neighbour in embedding space look up.
\fixM{
\cref{tab:examples} also\fixM{}{1.5} shows results when using $5$ and $10$ nearest neighbours.
Additionally,~\cref{fig:rc} shows results when using all of $N=1$ to $N=10$ nearest neighbours, as ``recall curves'' rather than tabulated.
}{}

Overall the best performing methods are \ref{B3} \textbf{Open-RadVLAD}~\cite{gadd2024open},
\ref{B4} \textbf{ResNet18-NetVLAD}, and \ref{B2} \textbf{RaPlace}~\cite{jang2023raplace}.

\fixM{\cref{fig:rc} shows the \texttt{Recall@N} curves corresponding to \cref{tab:examples}, averaged over all pairs of trajectories.
From this we can see that \ref{B3} \textbf{Open-RadVLAD} and \ref{B4} \textbf{ResNet18-NetVLAD} perform best with \ref{B3} \textbf{Open-RadVLAD} slightly better at lower recalled candidates (and thus perhaps being better suited to efficient deployment) but \ref{B4} \textbf{ResNet18-NetVLAD} performing slightly better when more candidates are recalled.
\ref{B2} \textbf{RaPlace} is worse than either of these at fewer recalled candidates but approaches \ref{B3} \textbf{Open-RadVLAD} when more database candidates are retrieved.}{4.5.7}

\fixM{\subsection{Failure cases}}{}
\fixM{}{4.5.5}

There is in fact a precipitous drop in localisation success rate for all methods in \fixM{}{4.4.4}\ref{DA1}-vs-\ref{DA2} in \texttt{twolochs}.
We attribute this to \texttt{twolochs} (\cref{sec:twolochs}) being the longest route (there therefore being more candidate scans to match to -- i.e. the search space being larger).
In~\cref{fig:twolochs_overview} this can be seen in the ground truth matrix (\textit{Top Right}) by the large region to the right of the white strip, where the query trajectory ends at \textit{Loch Laggan} and therefore only overlaps with the reference trajectory (\textit{Middle Right} in~\cref{fig:twolochs_overview}) for approximately \SI{40}{\percent} of the route, with the reference trajectory continuing north-east all the way along both bodies of water in \textit{Loch Laggan}.
This may also be due to the long periods driven along either \textit{Loch Laggan} or \textit{Lochan na h-Earba} where there is a dearth of distinctive scenery in an entire hemisphere of the radar scan.
An example scan from this type of scene can be seen in the third Cartesian frame in~\cref{fig:twolochs_overview} (\textit{Bottom}).
Therefore, \texttt{twolochs} is a useful test setting for robust recognition with vast maps as well as with featureless scenery, as radar place recognition matures.
For this challenging route, only radar-specific methods perform well and we have~\ref{B2} \textbf{RaPlace} performing best, with \ref{B3} \textbf{Open-RadVLAD} matching it in \texttt{Recall@10}.
\ref{B4} \textbf{ResNet18-NetVLAD} performs poorly on this pair of trajectories, likely due to clustering of the reference trajectory features, many of which are irrelevant to the query trajectory.

\fixS{
\fixM{}{4.5.6}Fig. 7 shows the \texttt{Recall@N} curves corresponding to Tab. III, averaged over all pairs of trajectories.
From this we can see that \ref{B3} \textbf{Open-RadVLAD} and \ref{B4} \textbf{ResNet18-NetVLAD} perform best with \ref{B3} \textbf{Open-RadVLAD} slightly better at lower recalled candidates (and thus perhaps being better suited to efficient deployment) but \ref{B4} \textbf{ResNet18-NetVLAD} performing slightly better when more candidates are recalled.
\ref{B2} \textbf{RaPlace} is worse than either of these at fewer recalled candidates but approaches \ref{B3} \textbf{Open-RadVLAD} when more database candidates are retrieved.}{}

\begin{figure}
\centering
\includegraphics[width=\columnwidth]{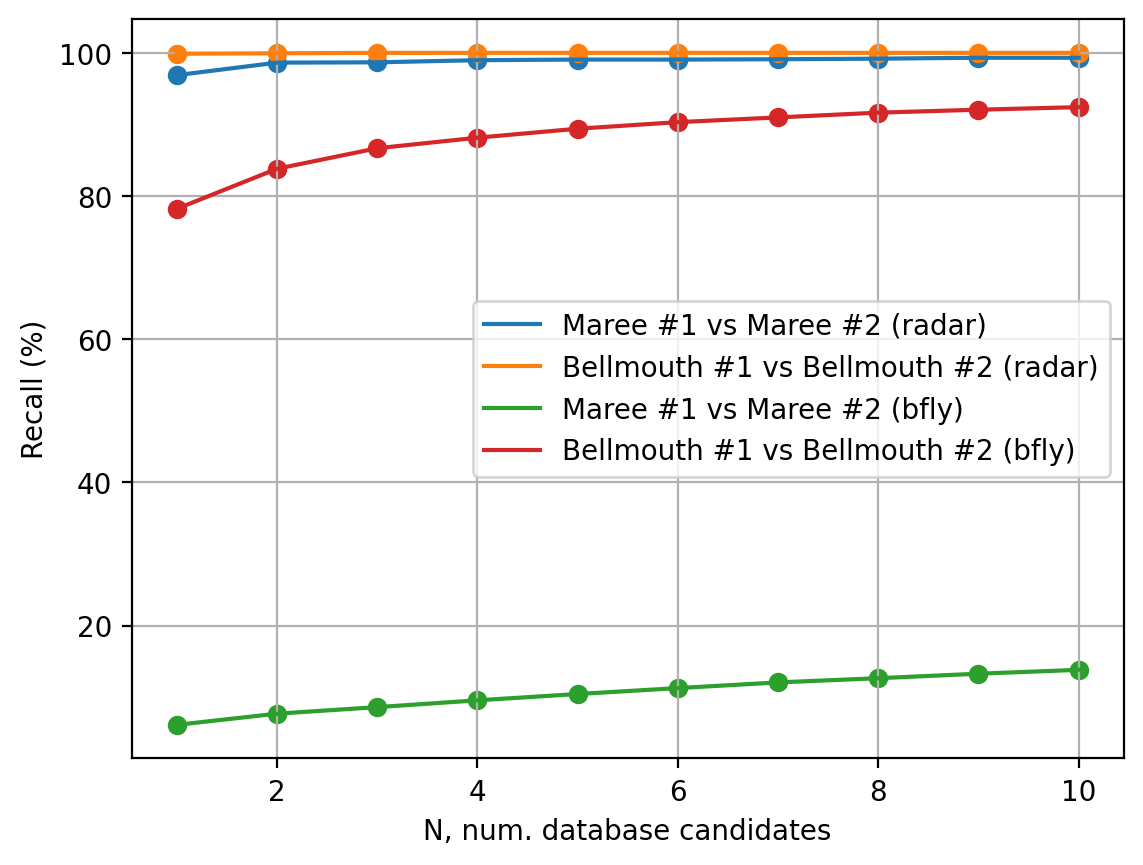}
\caption{\fixM{
\texttt{Recall@N} curves for four localisation experiments, between \ref{DB1} and \ref{DB2} (orange and red) as well as \ref{DM1} and \ref{DM2} (blue and green).
For radar (blue and orange) we test \ref{B3} while for vision (green and red) we test \ref{B8}.
}{3.2.1}}
\label{fig:bfly_rc}
\end{figure}

\fixM{\fixM{}{4.5.8}\subsection{Comparison against visual place recognition}}{}

\fixM{
Finally, \cref{fig:bfly_rc} shows a comparison between visual place recognition and radar place recognition on this dataset.
We see that radar has better performance in this task on this dataset across both experimental settings.
For example, we have almost \SI{100}{\percent} recall at any number of database candidates ($N$) -- see the orange curve -- while a vision-based approach has less than \SI{80}{\percent} \texttt{Recall@1} and this rises to only about \SI{90}{\percent} (red curve).
This was an experiment on \ref{DB1} versus \ref{DB2}, with little appearance variation (see~\cref{fig:BFLY}, top left)
The next experiment (blue and green) is on \ref{DM1} versus \ref{DM2}, which has a lot of appearance variation (see~\cref{fig:BFLY} where \ref{DM2} is under very poor lighting -- with driving taking place at night, in the dark).
Here, radar does only very slightly worse -- with over \SI{95}{\percent} \texttt{Recall@1} (\cref{fig:bfly_rc} leftmost blue dot and~\cref{tab:examples} third row, fourth column).
Performance with the camera, however, is drastically reduced -- e.g. dropping to approximately \SI{15}{\percent} \texttt{Recall@1}. 
This is an important result as it confirms that radar is somewhat immune to extreme appearance variation and is a useful counterpart (for localisation in this case) to the comparison to~\cite{sheeny2021radiate} where radar was shown to be quite robust to appearance change in urban environments in the task of vehicle\fixM{}{3.2.2} detection and tracking.
}{}

\section{Conclusion}
\label{sec:conclusion}

We have presented a novel radar dataset, carefully gathered under unique and challenging conditions.
This dataset has been crafted to catalyse advancements in the emerging field of radar place recognition.
We explored the use of our dataset in this task over a series of comprehensive experiments and evaluations, carried out across various open-source radar place recognition systems.
This was not only to showcase the dataset but also to establish a robust platform for future research in this evolving area.
\fixM{
We found that the best performing radar place recognition methods in terms of successful localisation queries over thousands of trials over $5$ pairs of sequences from $4$ distinct routes were the recently available open-source systems {OpenRadVLAD}, {RaPlace}, and {ResNet18-NetVLAD} the last of which is released alongside this dataset.
We also proved the usefulness of the dataset in confirming appearance-change robustness for radar in comparison to visual sensors.
}{1.6}

In the future, we would like to document new radar place recognition performance against this benchmark, and plan to do this online at e.g. \rurl{paperswithcode.com/datasets}.

\section*{Acknowledgements}

This work was supported by the Assuring Autonomy International Programme, a partnership between Lloyd’s Register Foundation and the University of York.
Matthew Gadd was supported by EPSRC Programme Grant ``From Sensing to Collaboration'' (EP/V000748/1).
Valentina Mu\cb{s}at was supported by a Google DeepMind Engineering Science Scholarship.
Efimia Panagiotaki was supported by a Google DeepMind Engineering Science Scholarship.
Lars Kunze was supported by EPSRC Project RAILS (EP/W011344/1).

We would also like to thank our partners at Navtech Radar, as well as Ian and Simon from \textit{Experience the Country} and Stuart and Tony from \textit{Highland All Terrain}, for expertly guiding and driving the vehicle around the data collection sites.

\bibliographystyle{IEEEtran}
\bibliography{biblio}


\begin{IEEEbiography}[{\includegraphics[width=1in,height=1.25in,clip,keepaspectratio]{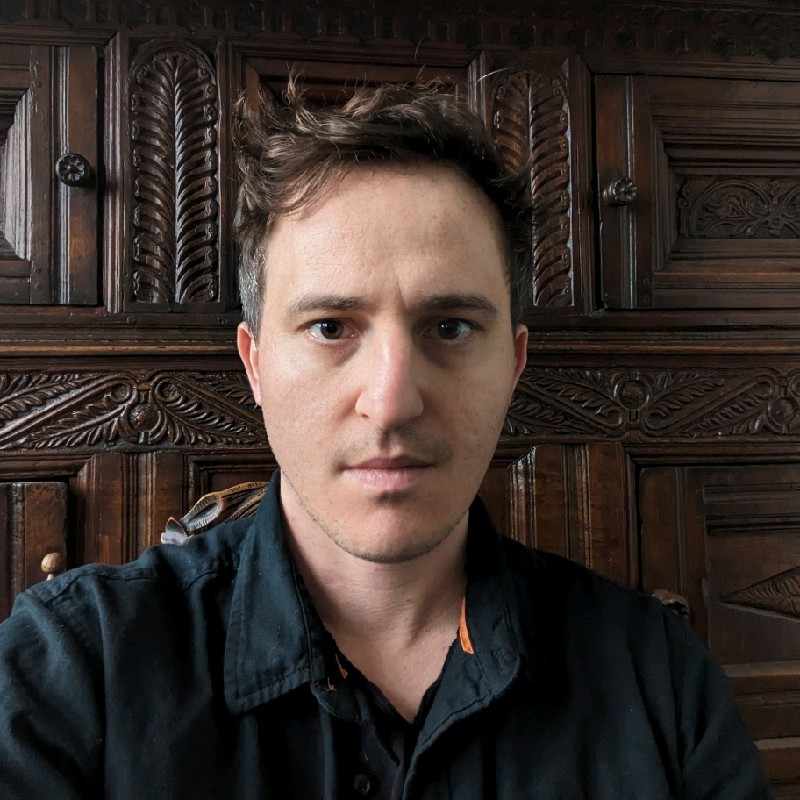}}]{
Matthew Gadd
}
is a Senior Research Associate in the Oxford Robotics Institute, Stipendiary Lecturer in Engineering Science at St Hilda's College, and Senior Software Engineer at Oxa.
He holds a DPhil (PhD) in Engineering Science from the University of Oxford
and a BSc in Mechatronics Engineering from the University of Cape Town.
\end{IEEEbiography}

\vskip -2\baselineskip plus -1fil\begin{IEEEbiography}[{\includegraphics[width=1in,height=1.25in,clip,keepaspectratio]{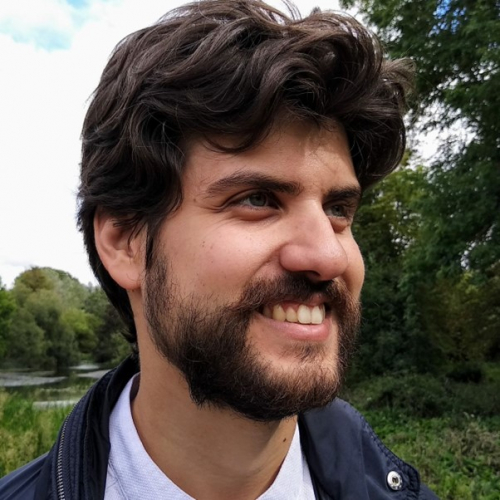}}]{
Daniele De Martini
}
is a Departmental Lecturer in Robotics at the University of Oxford and Stipendiary Lecturer in Engineering Science at Pembroke College.
He holds a PhD in Robotics from the University of Pavia, MSc in Mechatronic Engineering at the Politecnico di Torino, and BSc in Mechanical Engineering from the University of Pavia.
\end{IEEEbiography}

\vskip -2\baselineskip plus -1fil\begin{IEEEbiography}[{\includegraphics[width=1in,height=1.25in,clip,keepaspectratio]{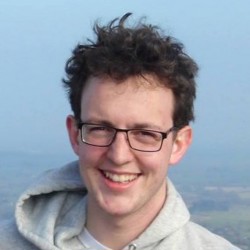}}]{
Oliver Bartlett
}
was the Trials and Media Coordinator at the Oxford Robotics Institute and is now a Fleet Readiness Engineer at Oxa.
He holds a MSc by Research as well as an MEng in Engineering Science from the University of Oxford.
\end{IEEEbiography}

\vskip -2\baselineskip plus -1fil\begin{IEEEbiography}
[{\includegraphics[width=1in,height=1.25in,clip,keepaspectratio]{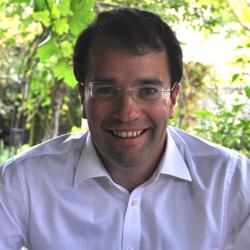}}]{
Paul Newman
}
is the BP Professor of Information Engineering at the University of Oxford and CTO at Oxa.
He holds a PhD in Autonomous Navigation from the Australian Centre for Field Robotics, University of Sydney and an MEng in Engineering Science from the University of Oxford.
\end{IEEEbiography}

\vskip -2\baselineskip plus -1fil\begin{IEEEbiography}[{\includegraphics[width=1in,height=1.25in,clip,keepaspectratio]{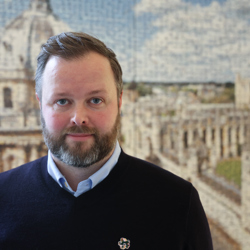}}]{
Lars Kunze
}
is a Full Professor in Safety for Robotics and Autonomous Systems at the Bristol Robotics Laboratory at UWE Bristol, Visiting Fellow in the Oxford Robotics Institute, and Stipendiary Lecturer in Computer Science at Christ Church College.
He holds a PhD (Dr. rer. nat.) from the Technical University of Munich, and an MSc in Computer Science and BSc in Cognitive Science from the University of Osnabr\"uck, Germany.
\end{IEEEbiography}

\end{document}